%% file: technical_report.tex
\documentclass[11pt]{fairmeta}
\let\ReportAuthorListBuilder\addtolist
\let\addtolist\ReportAuthorListBuilder
\usepackage{graphicx}  
\usepackage{fontawesome5} 
\input{math_commands.tex}

\usepackage{pifont}

\usepackage{hyperref}
\usepackage{url}
\usepackage{wrapfig}
\usepackage{placeins}
\usepackage[utf8]{inputenc} 
\usepackage[T1]{fontenc}    
\usepackage{hyperref}       
\usepackage{url}            
\usepackage{booktabs}       
\usepackage{amsfonts}       
\usepackage{nicefrac}       
\usepackage{microtype}      
\usepackage{xcolor}         
\usepackage{tikz}

\usepackage{booktabs}
\usepackage{enumitem}
\usepackage{rotating}
\usepackage[utf8]{inputenc}
\usepackage{amsmath,amssymb,amsfonts}

\usepackage{booktabs,bm}
\usepackage{listings}
\lstdefinestyle{recappython}{
 language=Python, basicstyle=\ttfamily\fontsize{9}{11}\selectfont,
 keywordstyle=\color{DeepPurple}\bfseries,
 commentstyle=\color{black!55}, stringstyle=\color{DeepPurple},
 numbers=left, numberstyle=\tiny\color{black!40}, numbersep=8pt,
 xleftmargin=18pt, xrightmargin=4pt,
 showstringspaces=false, keepspaces=true, columns=fullflexible,
 tabsize=4, breaklines=true, aboveskip=6pt, belowskip=6pt
}
\usepackage{colortbl}
\usepackage{multirow}
\usepackage{tabularx}
\usepackage{amsmath}       
\usepackage{amsfonts}      

\usepackage{xcolor}
\definecolor{commentcolor}{rgb}{0.5, 0.5, 0.5} 
\usepackage{booktabs}
\usepackage{adjustbox}
\definecolor{grey}{rgb}{0.89,0.71,0.57}
\definecolor{pink}{rgb}{1,0.94,1}
\definecolor{purple}{rgb}{0.84,0.78,1}
\definecolor{white}{rgb}{1,1,1}
\usepackage{amsmath}
\usepackage{mathtools}
\usepackage[utf8]{inputenc}
\usepackage{hyperref}
\usepackage{comment}
\usepackage{algorithm}
\usepackage{algorithmic}
\usepackage{listings}
\lstdefinestyle{recappython}{
 language=Python, basicstyle=\ttfamily\fontsize{9}{11}\selectfont,
 keywordstyle=\color{DeepPurple}\bfseries,
 commentstyle=\color{black!55}, stringstyle=\color{DeepPurple},
 numbers=left, numberstyle=\tiny\color{black!40}, numbersep=8pt,
 xleftmargin=18pt, xrightmargin=4pt,
 showstringspaces=false, keepspaces=true, columns=fullflexible,
 tabsize=4, breaklines=true, aboveskip=6pt, belowskip=6pt
}
\usepackage{wrapfig}
\usepackage{amsmath}
\usepackage{graphicx}
\usepackage{subcaption}
\usepackage{multirow}
\usepackage{booktabs}
\definecolor{Nuwa_blue}{rgb}{0,205,205}

\definecolor{rowblue}{RGB}{230, 240, 255}
\definecolor{rowgreen}{RGB}{230, 255, 230}
\definecolor{rowyellow}{RGB}{255, 255, 224}
\definecolor{rowpink}{RGB}{255, 230, 240}
\definecolor{rowcyan}{RGB}{224, 255, 255}
\definecolor{roworange}{RGB}{255, 240, 224}
\definecolor{rowviolet}{RGB}{240, 230, 255}
\definecolor{rowlime}{RGB}{240, 255, 224}
\definecolor{rowmint}{RGB}{224, 255, 240}
\definecolor{rowapricot}{RGB}{255, 224, 224}
\definecolor{rowgray}{RGB}{235, 235, 235}
\definecolor{DeepPurple}{RGB}{64, 20, 110}
\definecolor{LinkBlue}{RGB}{26, 77, 153}
\definecolor{metabg}{HTML}{F8FAFD}
\definecolor{metafg}{HTML}{202124}
\definecolor{metablue}{HTML}{1A73E8}

\hypersetup{
    colorlinks=true,
    pdftitle={Remembering What Matters for Long-Horizon Robotic World Models},
    pdfauthor={Hao Wu, Tianyu Shao, Shengju Qian, Yizhou Zhao, Fan Zhang, Penghao Zhao, Weiyan Wang, Yufei Huang, Fan Xu, Kun Wang, Xin Wang, Ming-Hsuan Yang, Yuxuan Liang},
    pdfsubject={Technical report},
    linkcolor=LinkBlue,
    citecolor=LinkBlue,
    filecolor=LinkBlue,      
    urlcolor=LinkBlue,
    }
\usepackage[utf8]{inputenc}
\usepackage{amsmath}
\usepackage{amssymb}
\usepackage{enumitem}
\usepackage{xcolor}
\usepackage{graphicx} 
\usepackage{tcolorbox} 
\tcbuselibrary{breakable,listings,skins}

\newlist{researchquestions}{enumerate}{1}
\setlist[researchquestions]{
    label=\textbf{RQ\arabic*:}, 
    leftmargin=*, 
    itemsep=8pt,
    font=\normalfont
}

\newtcolorbox{keyquestion}{
    colback=black!3,
    colframe=black!65,
    boxrule=0.7pt,
    arc=2mm,
    left=2mm,
    right=2mm,
    top=1.2mm,
    bottom=1.2mm,
    before skip=5pt,
    after skip=5pt
}

\usepackage[lining,tabular]{ebgaramond}
\usepackage{ebgaramond-maths}
\renewcommand{\sfdefault}{\rmdefault}
\DeclareTextFontCommand{\textbf}{\bfseries}
\renewcommand\authorformat[2][]{{\bfseries\mbox{#2$^{#1}$}}}
\renewcommand\bibfont{\fontsize{9.5}{11.2}\selectfont}
\newcommand{\method}{\textbf{ReCAP}}
\newcommand{\qwenjudge}{\mbox{\raisebox{-0.16em}{\includegraphics[height=1.05em]{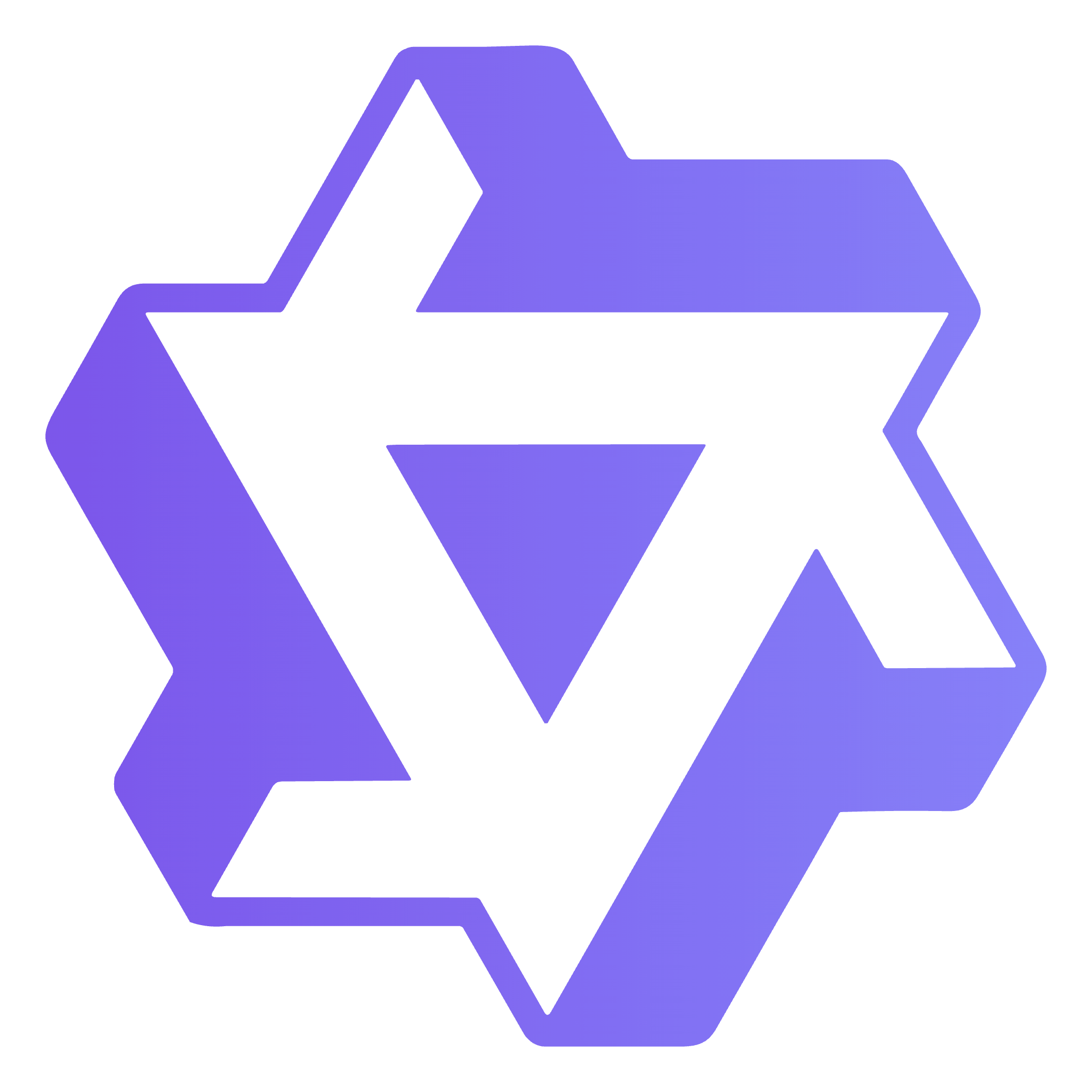}}\hspace{0.16em}Qwen3-VL}}
\newcommand{\internjudge}{\mbox{\raisebox{-0.16em}{\includegraphics[height=1.05em]{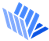}}\hspace{0.16em}InternVL3}}
\newcommand{\gptjudge}{\mbox{\raisebox{-0.16em}{\includegraphics[height=1.05em]{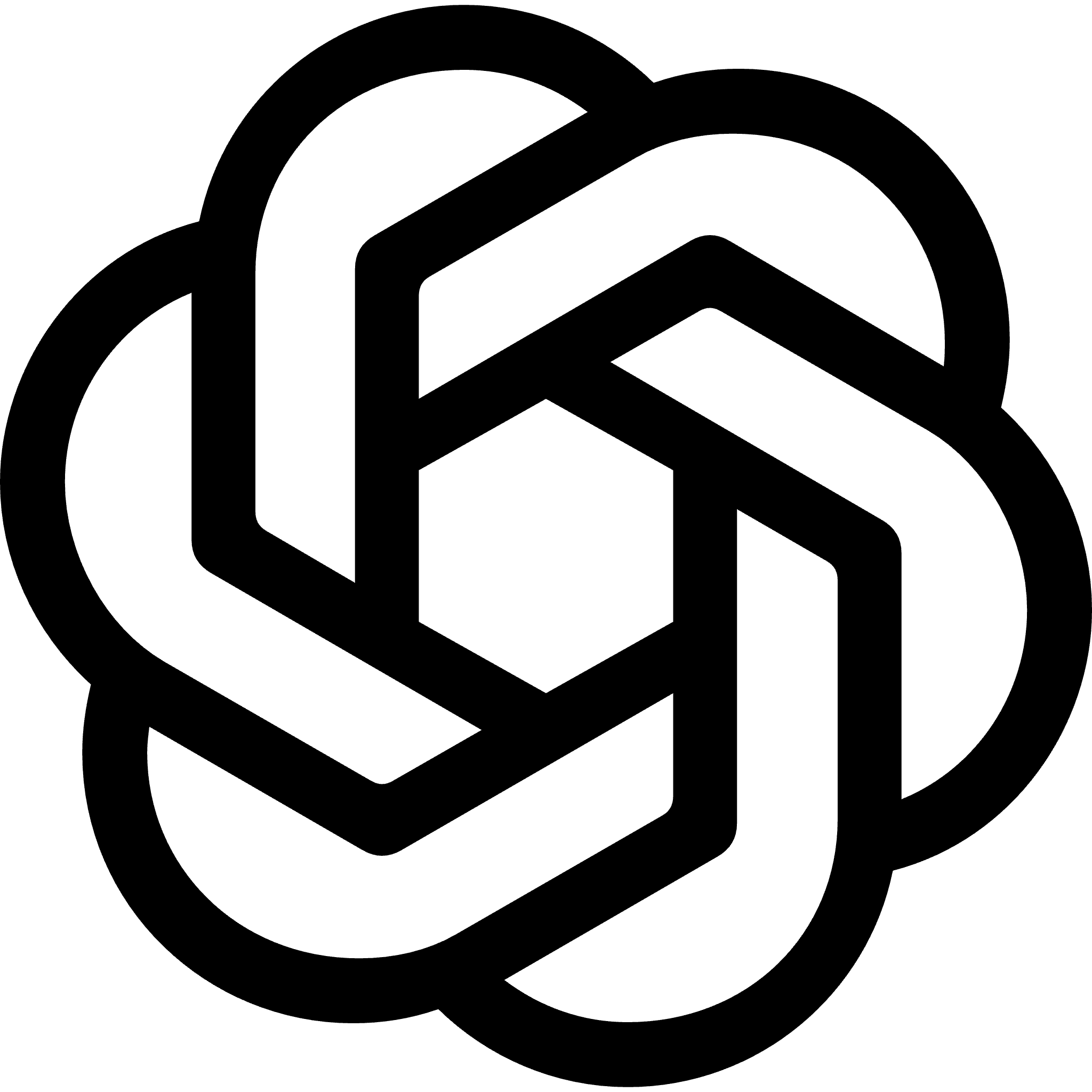}}\hspace{0.16em}GPT-5.5}}

\title{Remembering What Matters for Long-Horizon Robotic World Models}
\author[1]{Hao Wu}
\author[2]{Tianyu Shao}
\author[3]{Shengju Qian}
\author[4]{Yizhou Zhao}
\author[5]{Fan Zhang}
\author[6]{Penghao Zhao}
\author[6]{Weiyan Wang}
\author[6]{Yufei Huang}
\author[7]{Fan Xu}
\author[8]{Kun Wang}
\author[3]{Xin Wang}
\author[9,\dagger,\ddagger]{Ming-Hsuan Yang}
\author[1,\dagger]{Yuxuan Liang}
\affiliation[1]{The Hong Kong University of Science and Technology (Guangzhou)}
\affiliation[2]{North China Electric Power University}
\affiliation[3]{LIGHTSPEED}
\affiliation[4]{Carnegie Mellon University}
\affiliation[5]{The Chinese University of Hong Kong}
\affiliation[6]{Tencent}
\affiliation[7]{University of Science and Technology of China}
\affiliation[8]{Nanyang Technological University}
\affiliation[9]{University of California, Merced}
\contribution[\dagger]{Corresponding authors}
\contribution[\ddagger]{Project leader}
\abstract{\input{1_abstract}}
\metadata[{\faGithub\ Code}]{\href{https://github.com/Alexander-wu/ReCAP}{\nolinkurl{github.com/Alexander-wu/ReCAP}}}
\metadata[{\raisebox{-0.18em}{\includegraphics[height=1.1em]{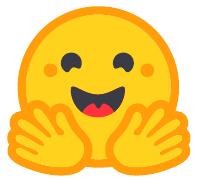}}\ Models}]{\href{https://huggingface.co/easylearning/recap-robot-models}{\nolinkurl{huggingface.co/easylearning/recap-robot-models}}}
\metadata[{\faGlobe\ Project Page}]{\href{https://alexander-wu.github.io/recap-project-page/}{\nolinkurl{alexander-wu.github.io/recap-project-page/}}}

\begin{document}

\maketitle
\input{2_introduction}
\input{3_related_work}
\input{4_method}
\input{5_experiment}

\input{6_conclusion}

\clearpage
\bibliography{references}
\bibliographystyle{assets/plainnat}
\clearpage
\beginappendix
\crefalias{section}{appendix}

\input{7_appendix}

\end{document}

%% file: math_commands.tex
\usepackage{amsmath,amsfonts,bm}

\def\eqref#1{equation~\ref{#1}}

\def\1{\bm{1}}

\DeclareMathAlphabet{\mathsfit}{\encodingdefault}{\sfdefault}{m}{sl}
\SetMathAlphabet{\mathsfit}{bold}{\encodingdefault}{\sfdefault}{bx}{n}



%% file: 1_abstract.tex
Autoregressive robot video world models based on discrete visual tokens remain
unreliable over long rollout horizons. Although these models typically retain
the entire generated history, a growing context does not necessarily improve
prediction quality. We investigate a possible \emph{memory-role conflict} in
Full Context decoding: persistent scene information, recent visual--action
states, and stale generated history compete for attention despite their
different predictive roles. This perspective motivates \method{}
(\underline{Re}ceding \underline{C}ontext with \underline{A}nchor
\underline{P}reservation), a training-free context-scheduling method for
long-horizon robot video world models. \method{} keeps the pretrained
tokenizer, world model, and visual decoder frozen. At inference time, it
preserves the initial scene anchor and a contiguous window of recent complete
visual--action blocks while evicting older generated history. The schedule
preserves within-block structure and temporal order, requires no additional
modules or parameter updates, and bounds active-context growth independently
of rollout length. We evaluate \method{} on RT-1, BridgeV2, LIBERO-90, and
CALVIN using frozen autoregressive world models. Across the evaluated
environments and rollout horizons, \method{} improves visual fidelity,
action-conditioned consistency, and robot structure preservation relative to
Full Context. On BridgeV2, it reduces LPIPS by $63.3\%$. In the system
benchmark at $H=64$, it reduces the logical KV footprint by $73.3\%$ and p50
latency by $6.51\%$. Attention analyses show increased attention to retained
recent states, while history interventions associate larger amounts of stale
context with higher prediction error. These observations are consistent with
the memory-role-conflict hypothesis, although they do not isolate it from
context-length and position-handling effects. Our results support inference-time
context scheduling as a practical approach to improving long-horizon robot
video prediction and computational efficiency without retraining.

%% file: 2_introduction.tex
\section{Introduction}
\label{sec:introduction}

\begin{figure*}[!t]
    \centering
    \includegraphics[width=\textwidth]{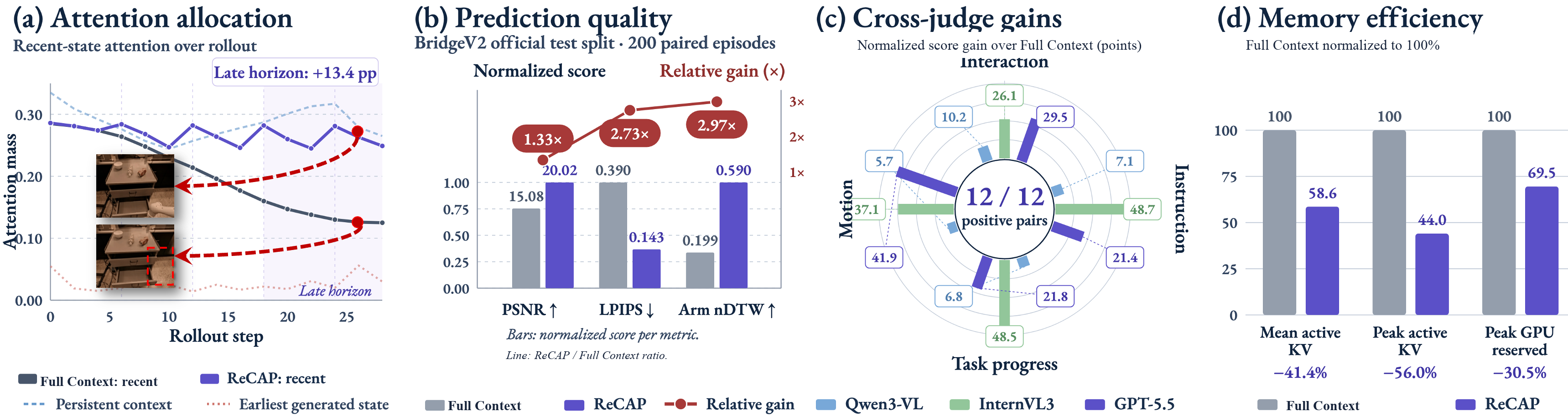}
    
    \caption{
    \textbf{\textit{\textnormal{\method{}} improves long-horizon prediction and memory efficiency.}}
    (a) \method{} preserves attention to recent visual states.
    (b) It improves prediction quality on BridgeV2.
    (c) Three VLM judges score \method{} higher across semantic criteria.
    (d) It reduces KV-cache and GPU-memory usage.
    Panels use dedicated samples and extrapolation horizons and are therefore not directly comparable to the fixed-horizon evaluations in Section~\ref{sec:experiments}.
    }
    \label{fig:intro}
\end{figure*}

Robot video world models predict how visual observations evolve under actions, providing a reusable substrate for planning, policy evaluation, and synthetic experience generation~\citep{finn2016unsupervised,ebert2018visualforesight,ha2018worldmodels,hafner2023dreamerv3}. Recent progress in discrete visual representation learning~\citep{oord2017neural,mentzer2024finite} and autoregressive sequence modeling~\citep{vaswani2017attention,yan2021videogpt} has made it possible to express observations and actions within a unified causal token sequence. This paradigm scales naturally from game environments~\citep{micheli2023iris,bruce2024genie} to interactive and robotic worlds~\citep{wu2024ivideogpt,ren2025videoworld,wu2025rlvr,wu2026roboalign}. Yet robot world models must do more than generate plausible short clips: they must preserve scene identity, robot morphology, object state, and action responsiveness over long horizons. These properties often deteriorate during autoregressive generation, limiting world models for long-horizon reasoning and control.

Most token-based video world models use \emph{full-history decoding}, which we denote as \emph{Full Context}: the initial context, action conditions, and every generated visual state remain active throughout the rollout~\citep{yan2021videogpt,bruce2024genie,wu2024ivideogpt}. This strategy appears natural because additional history should, in principle, provide more evidence for future prediction. Our experiments reveal the opposite behavior. As shown by the gray curve in Fig.~\ref{fig:intro}(a), the attention assigned to the most recent visual state steadily decreases as the rollout grows, even though distant context and early generated states remain accessible. The resulting predictions are initially plausible but progressively exhibit structural omission, weakened action response, and motion drift. Meanwhile, the active context and KV cache grow monotonically with the rollout horizon. This leads to the central question:

\raggedbottom
\begin{keyquestion}
\textbf{Key question:} \textit{How should an autoregressive robot video world model organize its inference context to remain dynamically informative and memory efficient over long horizons?}
\end{keyquestion}

We hypothesize that this failure partly reflects the different roles of
rollout history: persistent scene context, recent visual--action states,
and older generated history. Full Context places these tokens in a shared
attention pool; as history grows, the relative attention to recent states
can decline. We call this possible interference \emph{memory-role conflict}.
The traces in Fig.~\ref{fig:intro}(a) motivate this interpretation, but do
not by themselves establish causality. Attention sinks and streaming
inference have been studied in language models~\citep{xiao2024streamingllm,gu2025attentionsink,barbero2025llms};
robot video prediction additionally requires preserving the structure and
alignment of visual--action blocks.

Existing long-context and KV-cache methods primarily identify important individual tokens or compress a language-model cache~\citep{zhang2023h2o,liu2023scissorhands,ge2024fastgen,li2024snapkv}. Directly transferring token-level eviction to robot video rollouts is non-trivial because a physical time step is represented by a structured block of visual and action tokens; fragmenting this block can disrupt spatial coherence and causal alignment. Recent long-context video models also recognize that nearby frames and distant context play different roles~\citep{gu2025longcontext}, but typically address this distinction through architectural design or training. We instead ask whether the inference context of an already trained world model can be reorganized without modifying its parameters. This shift turns long-horizon rollout from a problem of retaining \emph{more} history into one of retaining the \emph{right} history.

We introduce \method{} (\underline{Re}ceding \underline{C}ontext with
\underline{A}nchor \underline{P}reservation), a training-free context scheduler.
It preserves the initial anchor and a contiguous window of recent complete
visual--action blocks, evicting older generated blocks after each update.
This retains within-frame structure, action alignment, and temporal order
while bounding active context independently of rollout length. The tokenizer,
world model, and decoder remain frozen. Unlike methods requiring long-context
training or architectural changes~\citep{gu2025longcontext}, \method{} changes
only the inference schedule.

Figure~\ref{fig:intro} summarizes the resulting gains under its dedicated overview protocols. Context reorganization repeatedly restores the influence of recent visual states, yielding a $13.4$-percentage-point advantage over Full Context at the late extrapolation horizon (Fig.~\ref{fig:intro}a). On its 200-pair BridgeV2 evaluation split, \method{} improves perceptual fidelity, pixel quality, and robot-trajectory consistency, reducing LPIPS by $63.3\%$ (Fig.~\ref{fig:intro}b). On 200 paired RT-1 episodes, \qwenjudge{}, \internjudge{}, and \gptjudge{} yield $12/12$ positive judge--criterion score differences (Fig.~\ref{fig:intro}c; Appendix~\ref{app:vlm_protocol}). Under the corresponding efficiency profile, \method{} reduces mean and peak active KV-cache usage by $41.4\%$ and $56.0\%$, respectively, and lowers peak reserved GPU memory by $30.5\%$ (Fig.~\ref{fig:intro}d), with near-neutral wall-clock overhead. These overview results use different samples, extrapolation lengths, and aggregation rules from the fixed-horizon analyses in Section~\ref{sec:experiments}; their numerical values should therefore not be compared directly. Across RT-1, BridgeV2, LIBERO-90, and CALVIN, the main experiments nevertheless show that the improvements persist across environments and rollout horizons.

History interventions associate additional stale context with increased
prediction error, although length and position effects remain confounded.
A separate prefix-length study tests scene/history budget allocation
(Appendix~\ref{app:prefix_length}), while human comparison and rubric sensitivity
analyses assess the semantic evaluation (Appendices~\ref{app:human_vlm_agreement}--\ref{app:vlm_sensitivity}).
Our contributions are:

\textbf{\textit{1. Empirical diagnosis}.} We identify memory-role conflict as a recurring failure mode of Full Context decoding in long-horizon robot video world models, and connect it to recent-state attention decay, structural drift, and unbounded context growth.  \textbf{\textit{2. Training-free context scheduling}.} We introduce \method{}, which preserves a persistent context anchor and recent visual--action states while evicting stale generated history, without retraining or architectural modification.  \textbf{\textit{3. Multi-level validation}.} We demonstrate consistent improvements in prediction quality, semantic consistency, and memory efficiency across robot datasets, supported by controlled interventions, attention analysis, and context-component ablations.

\flushbottom

%% file: 3_related_work.tex
\section{Related Work}

\textbf{\textit{Robot Video World Models}.}
World models learn action-conditioned dynamics for planning, policy evaluation, and synthetic experience generation~\citep{ha2018worldmodels,hafner2023dreamerv3}. Early robot video models predict future observations in pixel space~\citep{finn2016unsupervised,ebert2018visualforesight}, whereas discrete tokenizers and autoregressive Transformers represent observations and actions in a unified causal stream~\citep{oord2017neural,yan2021videogpt}. Genie, iVideoGPT, and RoboDreamer extend this paradigm to interactive and robotic settings~\citep{bruce2024genie,wu2024ivideogpt,ren2025videoworld,zhou2024robodreamer}, with reinforcement learning and reward alignment further improving action consistency~\citep{wu2025rlvr,wu2026roboalign}. These studies mainly target representations, architectures, or training, leaving inference-time context organization largely unexplored.

\textbf{\textit{Long-Horizon Autoregressive Video Modeling}.}
Autoregressive video world models condition on generated states to support causal, flexible-length rollouts~\citep{yan2021videogpt,bruce2024genie,wu2024ivideogpt,ren2025videoworld}. Yet train--inference mismatch compounds errors over time, degrading scene identity, object structure, motion, and action responsiveness~\citep{bengio2015scheduled,huang2025selfforcing,zhang2025framepack}. Retaining complete history also increases attention and KV-cache costs without guaranteeing better predictions~\citep{gu2025longcontext,po2025longcontext}. Hierarchical context, context packing, and state-space memory balance local and distant information, but typically require specialized architectures or additional training~\citep{gu2025longcontext,zhang2025framepack}. In contrast, \method{} reorganizes the existing visual--action context of a frozen robot world model at inference time.

\textbf{\textit{Memory and Context Management}.}
Memory management spans three main directions. H2O, Scissorhands, FastGen, and SnapKV evict or compress language-model KV caches according to token importance~\citep{zhang2023h2o,liu2023scissorhands,ge2024fastgen,li2024snapkv}. Streaming and sliding-window methods preserve initial and recent tokens~\citep{xiao2024streamingllm}, while attention-sink analyses explain the role of early tokens~\citep{gu2025attentionsink,barbero2025llms}. Transformer-XL, Compressive Transformer, and Recurrent Memory Transformer instead propagate or compress hidden states across segments~\citep{dai2019transformerxl,rae2020compressive,bulatov2022recurrent}.

These approaches mainly operate on language tokens or activations. Robot rollouts instead require complete visual--action blocks, since fragmenting a physical step can disrupt spatial coherence and action alignment. Unlike streaming language models, \method{} preserves a semantic scene anchor together with recent dynamics and schedules complete blocks without architectural modification or additional training.

%% file: 4_method.tex
\section{Problem Formulation}

We consider an action-conditioned robot video world model that receives an
initial observation sequence $\mathbf{o}_{1:C}$ and a future action sequence
$\mathbf{a}_{C:C+T-1}$, and generates future observations
$\hat{\mathbf{o}}_{C+1:C+T}$. The visual tokenizer $\mathcal{E}_v$ and action
tokenizer $\mathcal{E}_a$ map each observation and action to
$\mathbf{z}_t=\mathcal{E}_v(\mathbf{o}_t)$ and
$\mathbf{u}_t=\mathcal{E}_a(\mathbf{a}_t)$, respectively. We concatenate the
aligned representations into a visual--action block
$B_t=[\mathbf{z}_t;\mathbf{u}_t]$. A context policy $\mathcal{S}$ constructs
the active inference context $\mathcal{C}_t=\mathcal{S}(B_{1:t})$ from the
history. The frozen autoregressive world model predicts the next visual
representation, and the visual decoder reconstructs the corresponding frame:
\begin{equation}
    \hat{\mathbf{z}}_{t+1}
    \sim p_{\theta}\!\left(\,\cdot\mid\mathcal{C}_t\right),
    \qquad
    \hat{\mathbf{o}}_{t+1}
    =\mathcal{D}\!\left(\hat{\mathbf{z}}_{t+1};C_0\right).
    \label{eq:world_model_prediction}
\end{equation}
We seek a context policy that preserves accurate long-horizon prediction under
a finite context budget while keeping all model parameters fixed:
\begin{equation}
    \mathcal{S}^{*}
    =
    \arg\min_{\mathcal{S}}
    \mathbb{E}\!\left[
        \sum_{t=C}^{C+T-1}
        \ell\!\left(\hat{\mathbf{o}}_{t+1},\mathbf{o}_{t+1}\right)
    \right],
    \qquad
    \mathrm{s.t.}\quad
    |\mathcal{C}_t|\le M,\quad \theta\ \mathrm{fixed}.
    \label{eq:context_policy_objective}
\end{equation}
Here, $C_0$ is the preserved visual context, $\ell$ measures prediction error,
and $M$ is the input-context budget, excluding the next frame's output tokens. This objective requires $\mathcal{S}$ to construct a compact and
informative context from the visual--action history, balancing long-horizon
prediction quality with inference efficiency.

\section{Methodology}

\paragraph{Overview.}

\begin{figure*}[t]
    \centering
    \includegraphics[width=\textwidth]{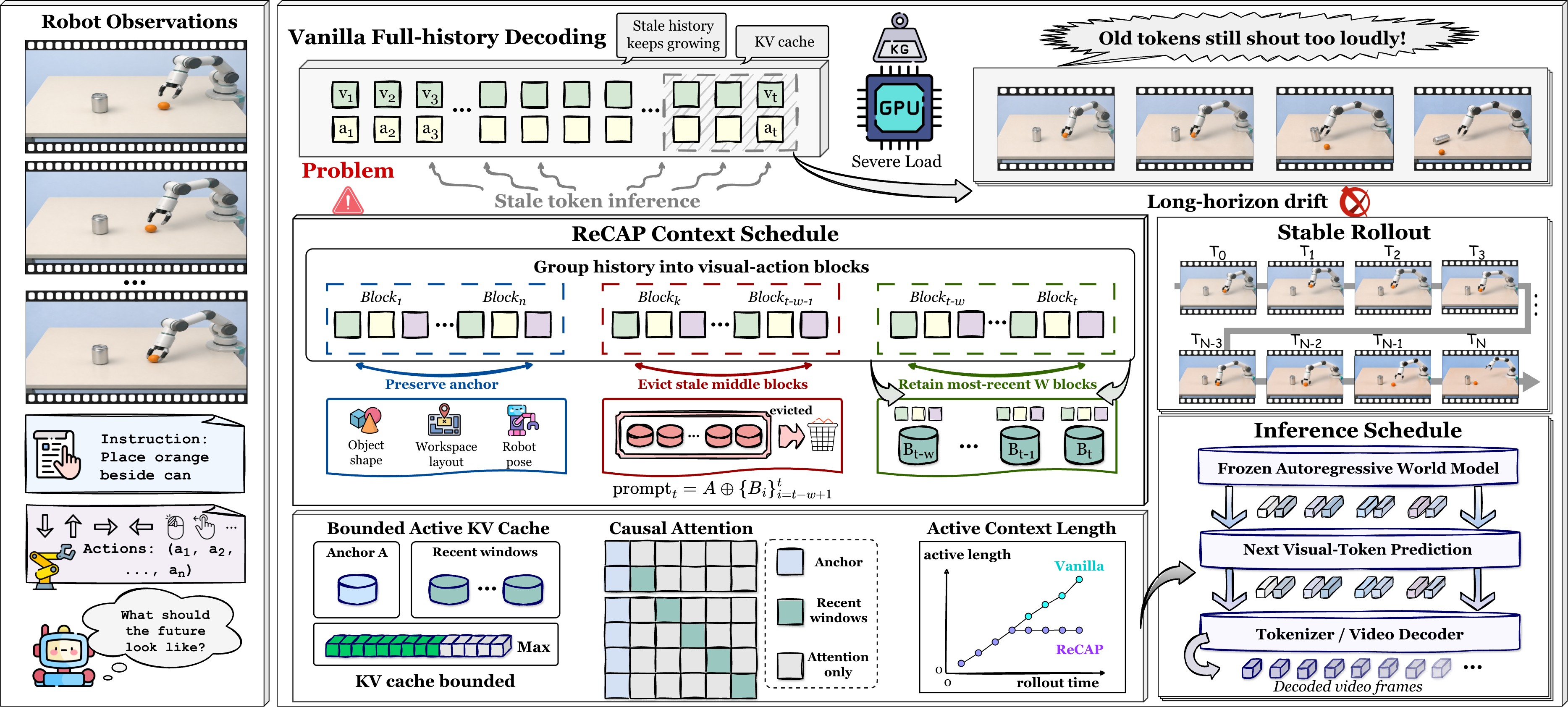}
    \caption{
    \textbf{\textit{Overview of \textnormal{\method{}}.}}
    Full Context retains all generated states, with growing context cost
    and increasing prediction drift in the evaluated rollouts.
    \method{} groups history into complete visual--action blocks, preserves the
    initial scene anchor and the most recent blocks, and evicts stale middle
    history. This training-free schedule keeps the active KV cache bounded while
    maintaining stable autoregressive rollouts.
    }
    \label{fig:recap_overview}
\end{figure*}

Figure~\ref{fig:recap_overview} illustrates \method{}, a training-free context
scheduler for pretrained robot video world models. At step $t$, Full Context
uses $P_t^{\mathrm{full}}=A\oplus B_{1:t}$, whereas \method{}
reorganizes this prompt by function: \ding{182} it preserves the scene anchor
$A$ from the real initial observations; \ding{183} it retains the most recent
$W$ complete visual--action blocks
$\mathcal{R}_t=B_{\max(1,t-W+1):t}$; and \ding{184} it evicts stale history
$\mathcal{O}_t=B_{1:t-W}$ from the active context and KV cache. It therefore
forms $P_t^{\mathrm{ReCAP}}=A\oplus\mathcal{R}_t$. The scheduler operates
blockwise in token space without changing model parameters or introducing
additional decoding and re-encoding.

\subsection{ReCAP Context Scheduling}

Given the real initial observations $\mathbf{o}_{1:C}$, we retain their scene
context $C_0$ together with the initial dynamics and action tokens as the fixed
anchor $A=C_0\oplus\mathbf{z}_0\oplus\mathbf{u}_0$. Its visual representations
come from real observations and contain no autoregressive prediction errors.
The scene context provides stable information about robot morphology, object
identity, and workspace layout. \method{} preserves $A$ throughout the rollout
as a persistent reference for future prediction.

At generation step $t$, the model has produced visual--action blocks
$B_{1:t}$, where $B_i=[\mathbf{z}_i;\mathbf{u}_i]$. \method{} defines the most
recent $W$ blocks as the recent trajectory
$\mathcal{R}_t=B_{\max(1,t-W+1):t}$ and the preceding blocks as stale history
$\mathcal{O}_t=B_{1:t-W}$; when $t\leq W$,
$\mathcal{O}_t=\varnothing$. The recent trajectory describes the current robot
state, object changes, and action evolution, whereas older generated history may offer diminishing predictive value
as its temporal distance grows. \method{} therefore removes $\mathcal{O}_t$ from the active
context and constructs
\begin{equation}
    P_t^{\mathrm{ReCAP}}
    =
    \underbrace{A}_{\text{persistent scene anchor}}
    \;\oplus\;
    \underbrace{
        B_{\max(1,t-W+1):t}
    }_{\mathcal{R}_t:\,\text{recent trajectory}},
    \qquad
    B_i \coloneqq [\mathbf{z}_i;\mathbf{u}_i].
    \label{eq:recap_prompt}
\end{equation}
where $\oplus$ concatenates token sequences in their original temporal order.
This context retains stable global scene information and locally relevant
dynamics while excluding generated blocks with indices $i\leq t-W$.
In the evaluated 1,931-token configuration, $A$ contains 1,280 scene tokens
plus an initial 93-token dynamics--action block, followed by $W=6$ recent
93-token blocks. Figure~\ref{fig:recap_token_sequence} in
Appendix~\ref{app:token_organization} shows this layout and the next-frame output.

\method{} updates the recent window online. After the model produces a new
block $B_{t+1}$, \method{} appends it to the current trajectory and retains only
the last $W$ blocks according to
$\mathcal{R}_{t+1}=\operatorname{Tail}_W(\mathcal{R}_t\oplus B_{t+1})$. Once
the window reaches its maximum size, each new block evicts the oldest block in
the window. This process schedules complete visual--action blocks and thus
preserves within-frame visual structure, state--action alignment, and the
original causal order. \method{} operates directly in token and KV-cache spaces
and never decodes generated tokens into images for re-encoding.

\subsection{Bounded-Context Inference}

At generation step $t$, \method{} uses
$P_t^{\mathrm{ReCAP}}=A\oplus\mathcal{R}_t$ as the active prompt and predicts
the next visual state with the original causal factorization:
\begin{equation}
    \hat{z}_{t+1,j}
    \sim
    p_\theta\!\left(
        \,\cdot\mid
        A,\mathcal{R}_t,\hat{\mathbf{z}}_{t+1,<j}
    \right),
    \qquad j=1,\ldots,n_v,
    \label{eq:recap_causal_generation}
\end{equation}
where $n_v$ denotes the number of visual tokens per observation. After
predicting all visual tokens, the model combines them with the aligned action
tokens to form $B_{t+1}=[\hat{\mathbf{z}}_{t+1};\mathbf{u}_{t+1}]$.

The cache retains entries associated with the selected anchor and recent
blocks. Let $\operatorname{KV}_t[X]$ denote entries for block $X$ in the active
cache at step $t$, including their dependence on preceding context:
\begin{equation}
    \operatorname{KV}_t^{\mathrm{ReCAP}}
    =\operatorname{KV}_t[A]
    \oplus\bigoplus_{i=\max(1,t-W+1)}^t\operatorname{KV}_t[B_i].
    \label{eq:recap_kv_cache}
\end{equation}
For backend $\beta$ with position indices $\pi_{t+1}^{(\beta)}$, the update is
\begin{equation}
    \operatorname{KV}_{t+1}^{\mathrm{ReCAP}}
    =\operatorname{Update}_{\beta}\!\left(
        \operatorname{KV}_t^{\mathrm{ReCAP}},
        P_{t+1}^{\mathrm{ReCAP}};\pi_{t+1}^{(\beta)}\right).
    \label{eq:recap_kv_update}
\end{equation}
Token selection alone does not imply unchanged cached activations. The BridgeV2
reference backend rebuilds KV from the retained prompt and rebases positions
after eviction (Appendix~\ref{app:bridge-reproducibility}); cache management
is separate from the block-selection algorithm.

Let $n_A$ and $n_B$ denote the token counts of the anchor and one
visual--action block, respectively. Full Context and \method{} use
\begin{equation}
    N_t^{\mathrm{full}}=n_A+t n_B,
    \qquad
    N_t^{\mathrm{ReCAP}}
    =n_A+\min(t,W)n_B
    \leq n_A+Wn_B.
    \label{eq:recap_context_bound}
\end{equation}
For a Transformer with $L$ layers, hidden dimension $d$, and $b$ bytes per
cached element, the corresponding KV-cache footprint satisfies
\begin{equation}
    M_{\mathrm{KV}}^{\mathrm{ReCAP}}(t)
    =
    2Ldb\,N_t^{\mathrm{ReCAP}}
    \leq
    2Ldb\left(n_A+Wn_B\right).
    \label{eq:recap_memory_bound}
\end{equation}
Once $t\geq W$, both the active prompt and KV cache remain bounded independently
of the total rollout length. \method{} preserves the original causal attention,
model parameters, and architecture, and introduces no pixel-space decoding or
re-encoding step. Algorithm~\ref{alg:recap_inference} provides the complete
inference procedure in the appendix.

%% file: 5_experiment.tex
\section{Experiments}
\label{sec:experiments}

\paragraph{Research questions.}
We organize our experiments around four research questions:
\begin{enumerate}[label=\textbf{\textit{RQ\arabic*:}}, leftmargin=*, itemsep=1.5pt, topsep=2pt, parsep=0pt]
    \item \textbf{Effectiveness and generality.} Does \method{} consistently improve prediction quality across datasets and evaluated backbones?
    \item \textbf{Long-horizon robustness.} How effectively does \method{} mitigate degradation over increasingly long autoregressive rollouts?
    \item \textbf{Mechanism and components.} Why does \method{} work, and how do the scene anchor, recent trajectory, and stale-history removal contribute?
    \item \textbf{Efficiency and scalability.} How does \method{} trade prediction quality against context size, KV-cache memory, and inference cost?
\end{enumerate}

\subsection{Experimental Setup}

\noindent\textcolor{DeepPurple}{\scriptsize$\blacktriangleright$}\hspace{0.35em}
\textbf{Datasets and backbones.}
We evaluate BridgeV2~\citep{walke2023bridgedata}, RT-1~\citep{brohan2023rt1},
LIBERO-90~\citep{liu2024libero}, and CALVIN~\citep{mees2022calvin} with
RLVR-World~\citep{wu2025rlvr}, iVideoGPT~\citep{wu2024ivideogpt}, and
RobotAlign-R1~\citep{wu2026roboalign}. Appendices~\ref{app:calvin-reproducibility}--\ref{app:bridge-reproducibility}
document dataset adaptation of the 127.1M RLVR-World transformer and its
tokenizer, including splits, action mappings, training curves, and checkpoint
selection. Adaptation precedes the scheduling comparison: each paired test
uses the same frozen checkpoint. ``Training-free'' refers to applying \method{}.

\noindent\textcolor{DeepPurple}{\scriptsize$\blacktriangleright$}\hspace{0.35em}
\textbf{Evaluation metrics.}
We evaluate generation quality from three complementary perspectives. PSNR, SSIM, and LPIPS measure visual fidelity. Arm Score, Shape IoU, and normalized dynamic time warping (nDTW) evaluate robot shape and trajectory consistency. Qwen3-VL~\citep{bai2025qwen3vl}, InternVL3~\citep{zhu2025internvl3}, and GPT-5.5~\citep{openai2026gpt55} further assess interaction quality, instruction following, task progress, and motion naturalness.

\noindent\textcolor{DeepPurple}{\scriptsize$\blacktriangleright$}\hspace{0.35em}
\textbf{Evaluation protocol and implementation.}
Paired rollouts share initial observations, actions, and generation seeds.
For the dedicated 200-pair RT-1 VLM study, judges score the same 16 sampled
frames per video with a shared four-criterion rubric and hidden method
identities (Appendix~\ref{app:vlm_protocol}, including the complete prompt).
Horizon-AUC and component comparisons report paired bootstrap 95\% intervals.
The efficiency benchmark uses vLLM in BF16; the additional BridgeV2 study
uses a separate FP32 PyTorch/Transformers backend. Appendix~\ref{app:implementation}
details the hardware and implementation settings.

\subsection{Overall Performance and Generality (RQ1)}
\label{sec:rq1}

\begin{table*}[!ht]
\centering
\caption{\textcolor{black}{\textbf{\textit{\method{} versus Full Context across GPT-style AR robot video world models.}}}
Comparisons pair identical episodes, sampling seeds, and action sequences.
$\Delta$ is the mean paired improvement. RQ1 uses each dataset's maximum test
horizon; subsequent experiments use fixed horizons. Better values are in
\textbf{bold}.}

\label{tab:main_results}
\footnotesize
\definecolor{tableteal}{RGB}{232,226,245}
\renewcommand{\arraystretch}{1.12}
\setlength{\tabcolsep}{3.0pt}
\begin{adjustbox}{max width=\textwidth}
\begin{tabular}{l|ccc|ccc|ccc|ccc}
\toprule
\textbf{Context Strategy} &
\multicolumn{3}{c|}{\textbf{BridgeV2}} &
\multicolumn{3}{c|}{\textbf{RT-1}} &
\multicolumn{3}{c|}{\textbf{LIBERO-90}} &
\multicolumn{3}{c}{\textbf{CALVIN}} \\
& LPIPS$\downarrow$ & nDTW$\uparrow$ & VLM$\uparrow$
& LPIPS$\downarrow$ & nDTW$\uparrow$ & VLM$\uparrow$
& LPIPS$\downarrow$ & nDTW$\uparrow$ & VLM$\uparrow$
& LPIPS$\downarrow$ & nDTW$\uparrow$ & VLM$\uparrow$ \\
\midrule

\rowcolor{tableteal}
\multicolumn{13}{l}{\textit{RLVR-World}} \\
Full Context
& 0.390 & 0.199 & 64.8
& 0.251 & 0.276 & 28.1
& 0.269 & 0.551 & 58.9
& 0.287 & 0.528 & 56.4 \\
\textbf{\method{} (Ours)}
& \textbf{0.143} & \textbf{0.590} & \textbf{72.6}
& \textbf{0.201} & \textbf{0.558} & \textbf{71.9}
& \textbf{0.217} & \textbf{0.631} & \textbf{67.8}
& \textbf{0.234} & \textbf{0.607} & \textbf{65.1} \\
\rowcolor{gray!8}
\quad $\Delta$
& \phantom{0}$-$0.247 & $+$0.391 & $+$7.8
& \phantom{0}$-$0.050 & $+$0.282 & $+$43.8
& \phantom{0}$-$0.052 & $+$0.080 & $+$8.9
& \phantom{0}$-$0.053 & $+$0.079 & $+$8.7 \\
\midrule

\rowcolor{tableteal}
\multicolumn{13}{l}{\textit{iVideoGPT}} \\
Full Context
& 0.254 & 0.587 & 61.3
& 0.267 & 0.561 & 58.6
& 0.283 & 0.529 & 55.7
& 0.301 & 0.506 & 53.2 \\
\textbf{\method{} (Ours)}
& \textbf{0.203} & \textbf{0.665} & \textbf{70.1}
& \textbf{0.215} & \textbf{0.639} & \textbf{67.6}
& \textbf{0.231} & \textbf{0.609} & \textbf{64.5}
& \textbf{0.248} & \textbf{0.586} & \textbf{61.9} \\
\rowcolor{gray!8}
\quad $\Delta$
& \phantom{0}$-$0.051 & $+$0.078 & $+$8.8
& \phantom{0}$-$0.052 & $+$0.078 & $+$9.0
& \phantom{0}$-$0.052 & $+$0.080 & $+$8.8
& \phantom{0}$-$0.053 & $+$0.080 & $+$8.7 \\
\midrule

\rowcolor{tableteal}
\multicolumn{13}{l}{\textit{RobotAlign-R1}} \\
Full Context
& 0.226 & 0.634 & 67.2
& 0.239 & 0.608 & 64.5
& 0.255 & 0.576 & 61.8
& 0.272 & 0.554 & 59.1 \\
\textbf{\method{} (Ours)}
& \textbf{0.178} & \textbf{0.709} & \textbf{75.8}
& \textbf{0.190} & \textbf{0.683} & \textbf{73.1}
& \textbf{0.206} & \textbf{0.654} & \textbf{70.2}
& \textbf{0.222} & \textbf{0.634} & \textbf{67.4} \\
\rowcolor{gray!8}
\quad $\Delta$
& \phantom{0}$-$0.048 & $+$0.075 & $+$8.6
& \phantom{0}$-$0.049 & $+$0.075 & $+$8.6
& \phantom{0}$-$0.049 & $+$0.078 & $+$8.4
& \phantom{0}$-$0.050 & $+$0.080 & $+$8.3 \\
\bottomrule
\end{tabular}
\end{adjustbox}

\end{table*}

\begin{figure*}[!ht]
    \centering
    \includegraphics[width=\textwidth]{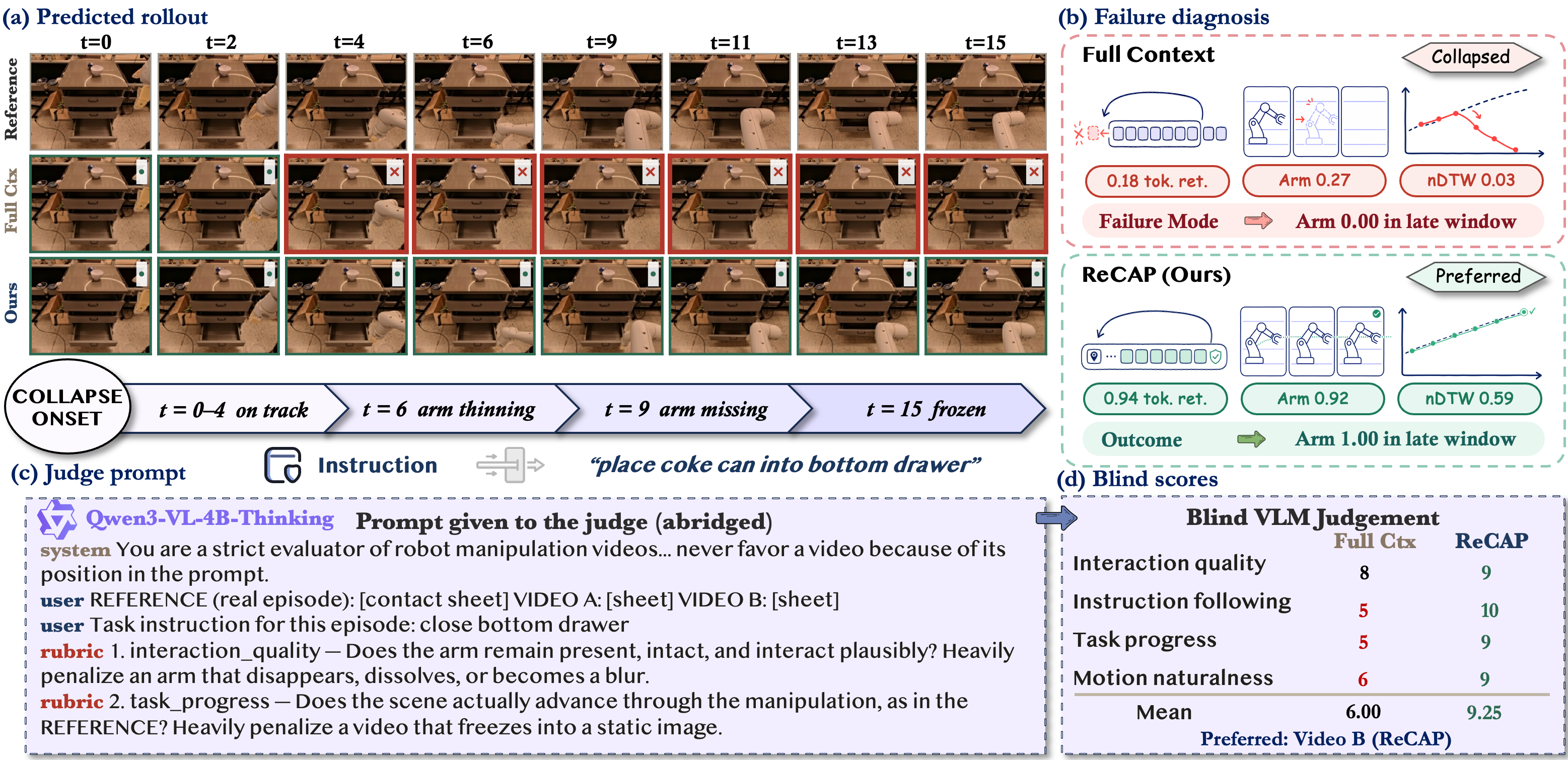}
    \caption{\textbf{\textit{Qualitative case study.}} \method{} preserves the robot arm
    throughout the rollout, whereas Full Context collapses; the blind VLM
    judge prefers \method{}. The displayed A/B prompt differs from the per-video
    RT-1 protocol in Appendix~\ref{app:vlm_protocol}.}
    \label{fig:rq1_case}
\end{figure*}

\textbf{Quantitative results.}
Table~\ref{tab:main_results} shows that \method{} outperforms Full Context in
all 36 dataset--backbone--metric combinations. With RLVR-World on BridgeV2,
\method{} reduces LPIPS from 0.390 to 0.143 and improves nDTW from 0.199 to
0.590. On RT-1, it increases nDTW and the VLM score by 0.282 and 43.8,
respectively. The consistent gains with iVideoGPT and RobotAlign-R1 further
demonstrate its generality across models and environments.
An additional 80-pair human comparison favors \method{} in 42 cases,
with 14 ties and 24 Full Context wins. Three-way VLM agreement with human
majorities is 66.3--72.5\% under the original protocol and 72.5--77.5\%
under the combined revision, whose tie rates approach the human rate.
The aggregate preference persists across tested variants, including removal
of the static penalty (Appendices~\ref{app:human_vlm_agreement}--\ref{app:vlm_sensitivity}).
These descriptive results provide limited human corroboration.

\textbf{Qualitative results.}
Figure~\ref{fig:rq1_case} qualitatively illustrates the long-horizon
improvement. Full Context begins to thin the robot arm at $t=6$, loses it at
$t=9$, and freezes by $t=15$, whereas \method{} preserves the robot structure
and task progress under the same initial observation, action sequence, and
sampling seed. Accordingly, the Arm Score increases from 0.27 to 0.92, nDTW
from 0.03 to 0.59, and the blind VLM mean score from 6.00 to 9.25.
Appendix~\ref{app:additional_qualitative} adds RT-1, CALVIN, and LIBERO-90
rollouts and selected high-gain BridgeV2 examples with 32 predicted frames;
these illustrate behavior rather than population-average gains.
\FloatBarrier

\subsection{Long-Horizon Robustness (RQ2)}
\label{sec:rq2}

We evaluate whether \method{} mitigates error accumulation during long
autoregressive rollouts. We generate 64 frames from identical initial
observations and ground-truth action sequences, and evaluate prediction quality
at $H=\{8,16,32,48,64\}$.

\begin{figure*}[!ht]
    \centering
    \includegraphics[width=\textwidth]{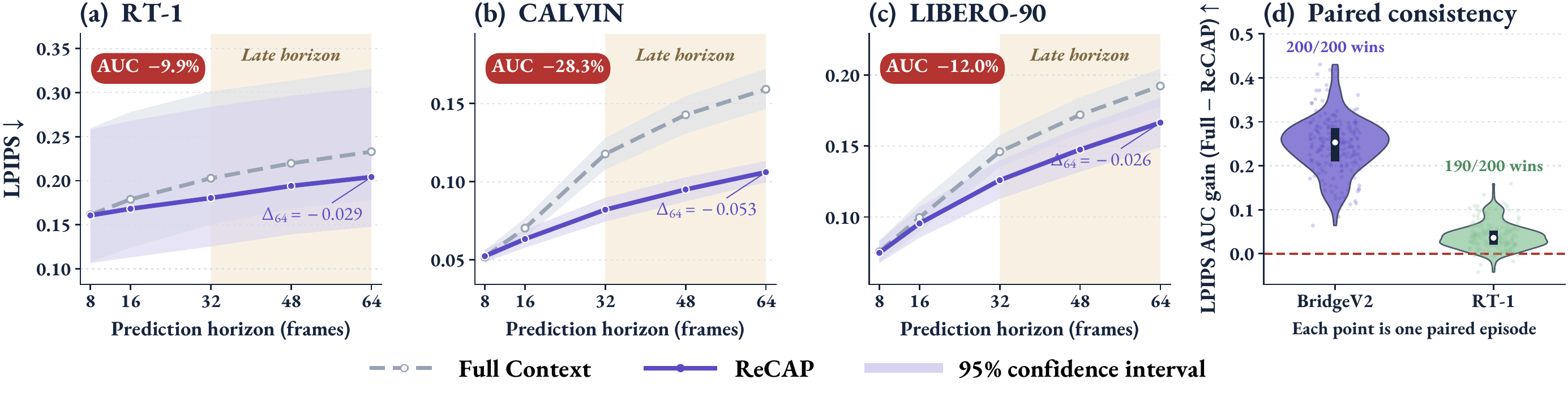}
    
    \caption{\textbf{\textit{\textnormal{\method{}} improves long-horizon robustness.}}
    (a--c) LPIPS across prediction horizons on three datasets; shaded bands
    denote 95\% confidence intervals. (d) Paired LPIPS AUC gains on independent
    BridgeV2 and RT-1 evaluation sets, where positive values favor \method{}.}
    
    \label{fig:rq2_multidataset_long_horizon}
\end{figure*}

\begin{table*}[!ht]
    \centering
    \caption{\textbf{\textit{Long-horizon prediction quality under paired rollouts.}}
    AUC gains are reported with 95\% confidence intervals; positive values favor
    \method{}. Results at $H=64$ are shown as Full Context $\rightarrow$
    \method{}, and win rates are reported for LPIPS/SSIM.}
    
    \label{tab:rq2_long_horizon}
    \footnotesize
    \definecolor{rqtwopurple}{RGB}{238,233,248}
    \renewcommand{\arraystretch}{1.22}
    \setlength{\tabcolsep}{5.2pt}
    \begin{adjustbox}{max width=\textwidth}
    \begin{tabular}{lccccc}
        \toprule
        \rowcolor{rqtwopurple}
        \textbf{Dataset}
        & \textbf{LPIPS AUC Gain $\uparrow$}
        & \textbf{SSIM AUC Gain $\uparrow$}
        & \textbf{$H{=}64$ LPIPS $\downarrow$}
        & \textbf{$H{=}64$ SSIM $\uparrow$}
        & \textbf{Win Rate} \\
        \midrule
        RT-1
        & $+0.0202\ [0.0136,\,0.0270]$
        & $+0.0207\ [0.0131,\,0.0286]$
        & $0.233 \rightarrow \mathbf{0.204}$
        & $0.703 \rightarrow \mathbf{0.732}$
        & $100\%/100\%$ \\
        \rowcolor{gray!5}
        CALVIN
        & $+0.0328\ [0.0245,\,0.0407]$
        & $+0.0222\ [0.0159,\,0.0287]$
        & $0.159 \rightarrow \mathbf{0.106}$
        & $0.833 \rightarrow \mathbf{0.867}$
        & $100\%/100\%$ \\
        LIBERO-90
        & $+0.0173\ [0.0120,\,0.0229]$
        & $+0.0192\ [0.0145,\,0.0241]$
        & $0.192 \rightarrow \mathbf{0.166}$
        & $0.798 \rightarrow \mathbf{0.826}$
        & $100\%/100\%$ \\
        \bottomrule
    \end{tabular}
    \end{adjustbox}
\end{table*}

\begin{figure*}[!ht]
    \centering
    \includegraphics[width=\textwidth]{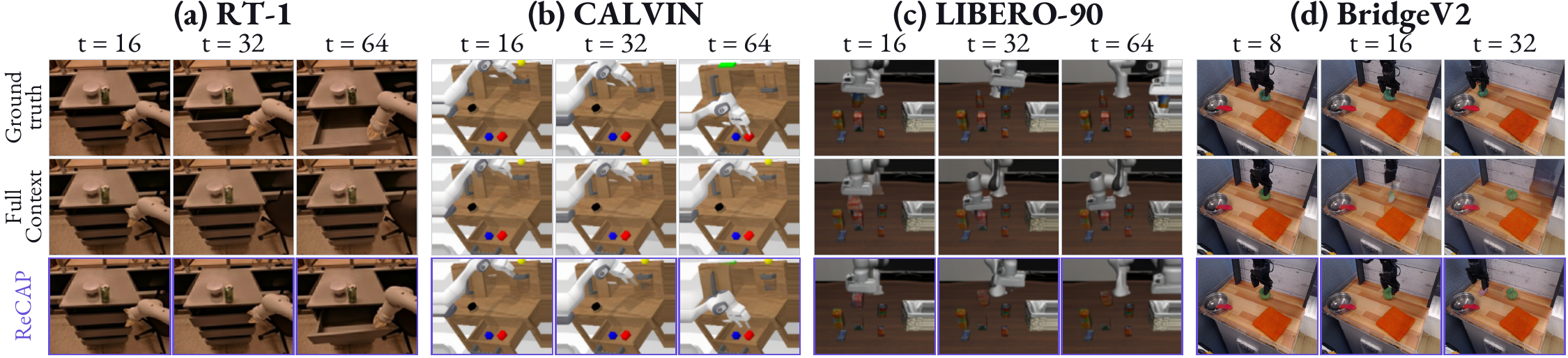}
    
    \caption{\textbf{\textit{Qualitative comparisons across four datasets.}}
    Selected cases from Appendix~\ref{app:additional_qualitative}, with three
    prediction times per dataset: $t=16,32,64$ for RT-1, CALVIN, and LIBERO-90;
    $t=8,16,32$ for the shorter BridgeV2 protocol. Rows show ground truth,
    Full Context, and \method{}.}
    \label{fig:rq2_long_horizon_cases}
\end{figure*}

\textbf{Quantitative results.}
Figure~\ref{fig:rq2_multidataset_long_horizon}(a--c) shows a small performance
gap at short horizons and an increasingly clear advantage for \method{} as the
rollout grows. \method{} reduces the LPIPS curve AUC by 9.9\%, 28.3\%, and
12.0\% on RT-1, CALVIN, and LIBERO-90, respectively. At $H=64$, it lowers
LPIPS from 0.233, 0.159, and 0.192 to 0.204, 0.106, and 0.166, while improving
SSIM on all three datasets. Table~\ref{tab:rq2_long_horizon} further shows that
every 95\% confidence interval excludes zero. Figure~\ref{fig:rq2_multidataset_long_horizon}(d)
uses separate BridgeV2 and RT-1 paired evaluation sets rather than the test sets
in panels (a--c), and its gains remain concentrated above zero.

\textbf{Qualitative results.}
Figure~\ref{fig:rq2_long_horizon_cases} compares matched rollouts on all four
datasets. These selected examples illustrate better preservation of robot
structure and scene evolution under \method{}. BridgeV2 uses the separate
32-frame protocol in Appendix~\ref{app:bridge-reproducibility}; the displayed
cases do not estimate population-average performance.
\subsection{Mechanism and Components (RQ3)}
\label{sec:rq3}

\begin{figure*}[!h]
    \centering
    \includegraphics[width=\textwidth]{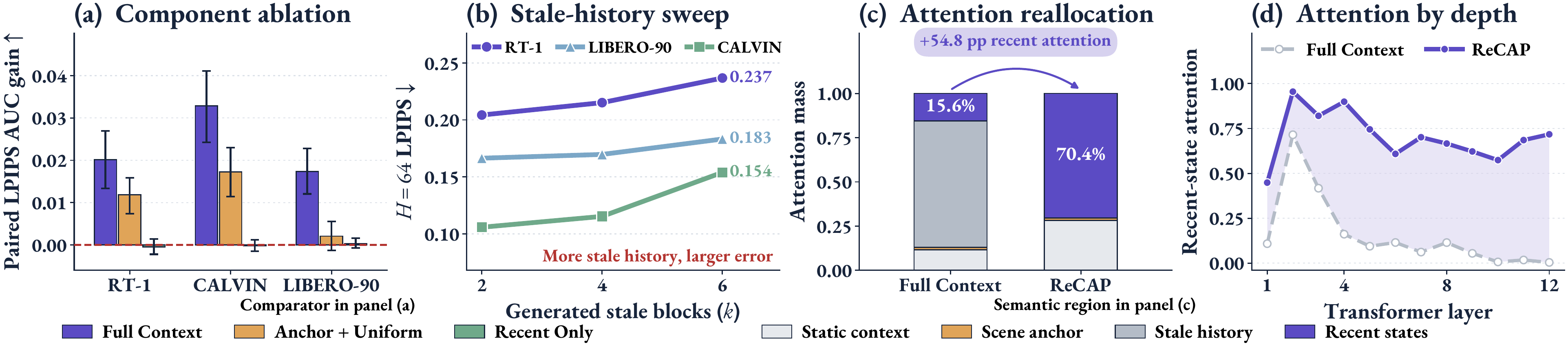}
    \caption{\textbf{\textit{Mechanism analysis of \textnormal{\method{}}.}}
    (a) Component ablation with 95\% bootstrap confidence intervals.
    (b) $H{=}64$ LPIPS after reinserting $k$ stale blocks.
    (c) Layer-averaged semantic attention and (d) recent-state attention by
    layer on CALVIN C06 at $H{=}64$.}
    \label{fig:rq3_mechanism}
\end{figure*}

At a fixed budget, \emph{Anchor + Uniform} retains the scene anchor and
uniformly samples history; \emph{Recent Only} retains recent states without
the anchor.

\textbf{Component ablation.}
In Figure~\ref{fig:rq3_mechanism}(a), paired LPIPS AUC improves over Full
Context by 0.0202, 0.0328, and 0.0173 on RT-1, CALVIN, and LIBERO-90, and over
Anchor + Uniform by 0.0118, 0.0173, and 0.0021. This favors contiguous recent
states over uniform history sampling. Gains over Recent Only remain within
$\pm0.0005$, with confidence intervals including zero: the anchor adds no
measurable LPIPS AUC benefit in this comparison.

\textbf{Scene-prefix allocation.}
Using the final 20k LIBERO-90 checkpoint on 450 held-out episodes and five
seeds, expanding the scene prefix from 4 to 1,280 tokens under a 1,931-token
cap shortens the recent window from 19 to 6 complete blocks. Late LPIPS
improves from 0.208 to 0.203 and nDTW from 0.619 to 0.628, with paired 95\% CIs
$[-0.007,-0.003]$ and $[+0.005,+0.013]$ for the differences.
This tests joint scene/history allocation using different metrics from the
AUC ablation above. Appendix~\ref{app:prefix_length} reports effective lengths
and all comparisons; the shared cap does not give identical token counts.

\textbf{Stale-history intervention.}
Reinserting $k=2$ to 6 stale blocks monotonically raises $H{=}64$ LPIPS
from 0.204 to 0.237 on RT-1, 0.106 to 0.154 on CALVIN, and 0.166 to 0.183
on LIBERO-90 (Fig.~\ref{fig:rq3_mechanism}b). Reinsertion also lengthens the
context, so it does not isolate stale content from length and position effects.

\textbf{Attention analysis.}
On CALVIN C06 at $H{=}64$, Full Context assigns 71.5\% of attention to stale
history and 15.6\% to recent states. Under \method{}, recent-state attention
is 70.4\%, rising in all 12 layers, while anchor attention changes from
1.37\% to 1.45\% (Fig.~\ref{fig:rq3_mechanism}c--d). Removing tokens also
changes attention normalization. Moreover, the 1,931-token inference budget
matches the adaptation sequence length (Appendix~\ref{app:implementation}).
These observations support the memory-role-conflict hypothesis but do not
separate it from length, position, and cache-update effects.

\FloatBarrier

\begin{table}[!htbp]
    \centering
    \begin{minipage}{0.52\textwidth}
    \centering
    \caption{\textbf{\textit{System efficiency summary.}}
    $H{=}64$ values are macro averages; throughput uses $C{=}256$.}
    
    \label{tab:rq4_compact}
    \fontsize{9pt}{10.8pt}\selectfont
    \definecolor{rqfourpurple}{RGB}{232,226,245}
    \renewcommand{\arraystretch}{1.08}
    \setlength{\tabcolsep}{1.8pt}
    \begin{tabularx}{\linewidth}{@{}>{\raggedright\arraybackslash}Xrrr@{}}
        \toprule
        \rowcolor{rqfourpurple}
        \textbf{Metric} & \textbf{Full} & \textbf{\method{}} & \textbf{Change} \\
        \midrule
        Tokens $\downarrow$ & 7,232 & \textbf{1,931} & \textbf{$-73.3\%$} \\
        \rowcolor{gray!5}
        Logical KV (MiB) $\downarrow$ & 254.25 & \textbf{67.89} & \textbf{$-73.3\%$} \\
        p50 (ms) $\downarrow$ & 312.95 & \textbf{292.57} & \textbf{$-6.51\%$} \\
        \rowcolor{gray!5}
        $C{=}256$ throughput $\uparrow$ & 1,596 & \textbf{6,834} & \textbf{$4.28\times$} \\
        \bottomrule
    \end{tabularx}
    
    \end{minipage}
\end{table}

\subsection{Efficiency and Scalability (RQ4)}
\label{sec:rq4}

\begin{figure*}[!h]
    \centering
    \includegraphics[width=\textwidth]{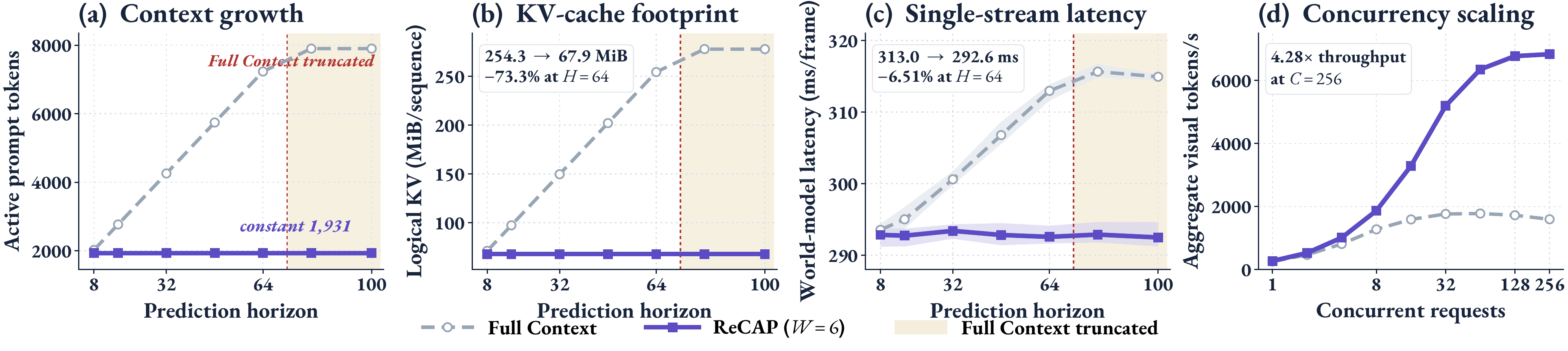}
    
    \caption{\textbf{\textit{\textnormal{\method{}} bounds context cost and
    improves concurrent throughput.}}
    (a) Prompt tokens, (b) logical BF16 KV, (c) p50 world-model latency, and
    (d) concurrent throughput. Shading marks Full Context truncation; bands in
    (c) show the range across three datasets.}
    \label{fig:rq4_efficiency}
\end{figure*}

\textbf{Efficiency protocol.}
An exclusive H20 runs a persistent vLLM 0.13.0 BF16 engine in
eager/\texttt{-O0} mode, with \texttt{max\_model\_len}=8,192 and 80 output
tokens reserved per frame. \method{} uses $W{=}6$. Automatic prefix caching
is disabled here and evaluated separately. Each single-stream measurement
uses two warm-up and five measured runs with tokenizer-produced prompts.
Latency covers world-model generation only; logical KV excludes allocator
overhead, activations, and workspace.

\textbf{Bounded cost.}
At $H{=}64$, \method{} reduces the active prompt from 7,232 to 1,931 tokens
and logical KV from 254.25 to 67.89\,MiB, a 73.3\% reduction
(Table~\ref{tab:rq4_compact}). It also lowers p50 latency from 312.95 to
292.57\,ms/frame ($-6.51\%$). Beginning after rollout step 72, Full Context is
truncated to at most 7,900 prompt tokens; at $H{=}100$, \method{} remains
bounded at 1,931 tokens
(Figure~\ref{fig:rq4_efficiency}(a--c)).

\textbf{Concurrent scaling.}
The throughput advantage grows from $1.07\times$ at one request to
$2.94\times$ at 32 requests and $4.28\times$ at 256 requests
(Figure~\ref{fig:rq4_efficiency}(d)). Neither method fails within the tested
range; thus, 256 denotes the largest tested concurrency rather than a hard
capacity limit.

%% file: 6_conclusion.tex
\raggedbottom
\begin{samepage}
\section{Conclusion}

We presented \method{}, a training-free context scheduler that preserves the
initial anchor and recent complete visual--action blocks. Across evaluated
datasets and backbones, it improves long-horizon prediction over Full Context.
Prefix-allocation and human--VLM analyses provide complementary evidence under
their respective protocols. History and attention analyses are consistent
with memory-role conflict, while context-length and backend effects remain
unresolved. At $H=64$, \method{} reduces logical KV use by 73.3\% and achieves
$4.28\times$ throughput at 256 concurrent requests. These results support
context scheduling for efficient robot video prediction with frozen models.
Context composition thus merits attention alongside model capacity in
long-horizon prediction. Future work should isolate content, position, and
cache effects through matched controls. Adaptive context allocation and
closed-loop robot evaluation would further clarify when this schedule is
most useful.\par
\end{samepage}
\clearpage
\flushbottom

%% file: 7_appendix.tex
\section{ReCAP Inference Algorithm}
\label{app:recap_algorithm}

\subsection{Token Organization}
\label{app:token_organization}

Figure~\ref{fig:recap_token_sequence} shows the evaluated 1,931-token input:
a 1,373-token fixed prefix $A_0$ (the anchor $A$ in Equation~\ref{eq:recap_prompt})
and six recent visual--action blocks. The next frame's 80 output tokens are
excluded from this input budget.

\input{Figures/recap_token_sequence_draft}

\clearpage
\subsection{Inference Procedure}
\label{app:inference_procedure}

Algorithm~\ref{alg:recap_inference} uses frozen model and decoder interfaces.
The \texttt{anchor} is the complete fixed prefix $A_0$, including the initial
action block; subsequent action blocks follow their aligned visual blocks.
The decoder uses the preserved visual context $C_0$.

\begin{algorithm}[H]
\caption{\method{} inference}
\label{alg:recap_inference}
\begin{lstlisting}[style=recappython]
def recap(anchor, actions, model, decoder, W, n_visual):
    # actions[t] follows predicted visual block t + 1.
    # anchor already includes the initial action block.
    assert W >= 1
    recent = []
    rollout = []

    for action in actions:
        # Keep the fixed anchor and recent complete blocks.
        prompt = list(anchor)
        for block in recent:
            prompt.extend(block)

        # Autoregressively sample one frame's visual tokens.
        visual = []
        for _ in range(n_visual):
            token = model.sample_next(prompt + visual)
            visual.append(token)

        rollout.append(decoder.decode(visual, context=anchor))
        block = visual + list(action)
        recent.append(block)
        recent = recent[-W:]  # Evict the oldest whole block.

    return rollout
\end{lstlisting}
\end{algorithm}

Predicted visual tokens enter the context directly; decoded frames are
outputs and are never re-encoded. The algorithm specifies token selection
and generation order. Backend-specific KV updates, including cache rebuilding
and position rebasing after eviction in the BridgeV2 implementation, are
described in Appendix~\ref{app:bridge-reproducibility}.

\section{Implementation and Systems Details}
\label{app:implementation}

\subsection{Hardware Environment}

Model execution uses a server with eight NVIDIA H20 GPUs (95.6\,GiB each)
and two AMD EPYC 9K84 96-core CPUs. The RQ4 systems benchmark reserves one
H20 exclusively; Ray handles job scheduling only.

\subsection{Software Environment}

\begin{table}[H]
    \centering
    \caption{\textbf{\textit{Software environment used for model execution and evaluation.}}
    Versions marked ``metrics'' refer to the separate metric-evaluation
    environment.}
    \label{tab:software_environment}
    \small
    \renewcommand{\arraystretch}{1.08}
    \setlength{\tabcolsep}{5pt}
    \begin{tabularx}{\columnwidth}{@{}lX@{}}
        \toprule
        \textbf{Component} & \textbf{Version} \\
        \midrule
        Python & 3.13.11 \\
        PyTorch / torchvision & 2.9.1 / 0.24.1+d801a34 \\
        CUDA (PyTorch build) / cuDNN & 12.9 / 9.1.6.0 \\
        NCCL & 2.27.3 \\
        Transformers & 4.57.1; 4.57.1.post1 (metrics) \\
        vLLM & 0.13.0 \\
        Ray & 2.51.1 (scheduling only) \\
        Accelerate & 1.9.0 \\
        PEFT & 0.16.0; 0.17.1 (metrics) \\
        safetensors / flashinfer & 0.7.0 / 0.5.3 \\
        lpips / piqa & 0.1.4 (VGG) / 1.3.2 \\
        NumPy / SciPy & 1.26.4 / 1.17.1 \\
        Pillow / imageio & 12.0.0 / 2.37.0 \\
        matplotlib & 3.11.1 (rendering only) \\
        \bottomrule
    \end{tabularx}
\end{table}

\subsection{World-Model Serving and Context Limits}

The adapted world models share the RLVR-World RT-1 architecture: 12 layers,
hidden size 768, 12 attention heads, feed-forward size 3,072, vocabulary
size 9,008, and maximum context length 8,192 (127.1M parameters).
For the efficiency benchmark, we use one persistent vLLM 0.13.0 engine in
BF16 with eager execution (optimization level \texttt{-O0}, CUDA graphs
disabled). We reserve 80 tokens for each frame's visual output, cap the
Full Context prompt at 7,900 tokens, and treat it as truncated after rollout
step 72.

For \method{}, the scene context contains 1,280 tokens and the fixed prefix
also retains the initial 93-token visual--action block
(Figure~\ref{fig:recap_token_sequence}). Each recent block contains 93 tokens,
and the recent window uses $W=6$. Its active prompt is consequently bounded by
\begin{equation}
    1280 + 93(1+W) = 1{,}931 \ \text{tokens}.
\end{equation}
Automatic prefix caching is disabled for the main efficiency results. We
evaluate the APC-enabled configuration separately as a deployment-sensitivity
experiment.

We report the analytical logical BF16 KV footprint rather than allocated GPU
memory. For the evaluated model, the per-token footprint is
\begin{equation}
    2 \times 12 \times 12 \times 64 \times 2
    = 36{,}864 \ \text{bytes/token},
\end{equation}
where the factors correspond to key/value tensors, Transformer layers, KV
heads, head dimension, and bytes per BF16 element. This calculation excludes
KV-block rounding, allocator metadata, activations, and runtime workspace.

\input{Appendix/appendix_calvin_reproducibility}
\input{Appendix/appendix_libero_reproducibility}
\clearpage
\input{Appendix/appendix_bridge_reproducibility}
\FloatBarrier
\input{Appendix/appendix_additional_qualitative}
\clearpage
\input{Appendix/appendix_vlm_protocol}
\clearpage
\input{Appendix/appendix_prefix_length}

%% file: Figures/recap_token_sequence_draft.tex
\begin{figure}[H]
  \centering
  \includegraphics[width=\textwidth]{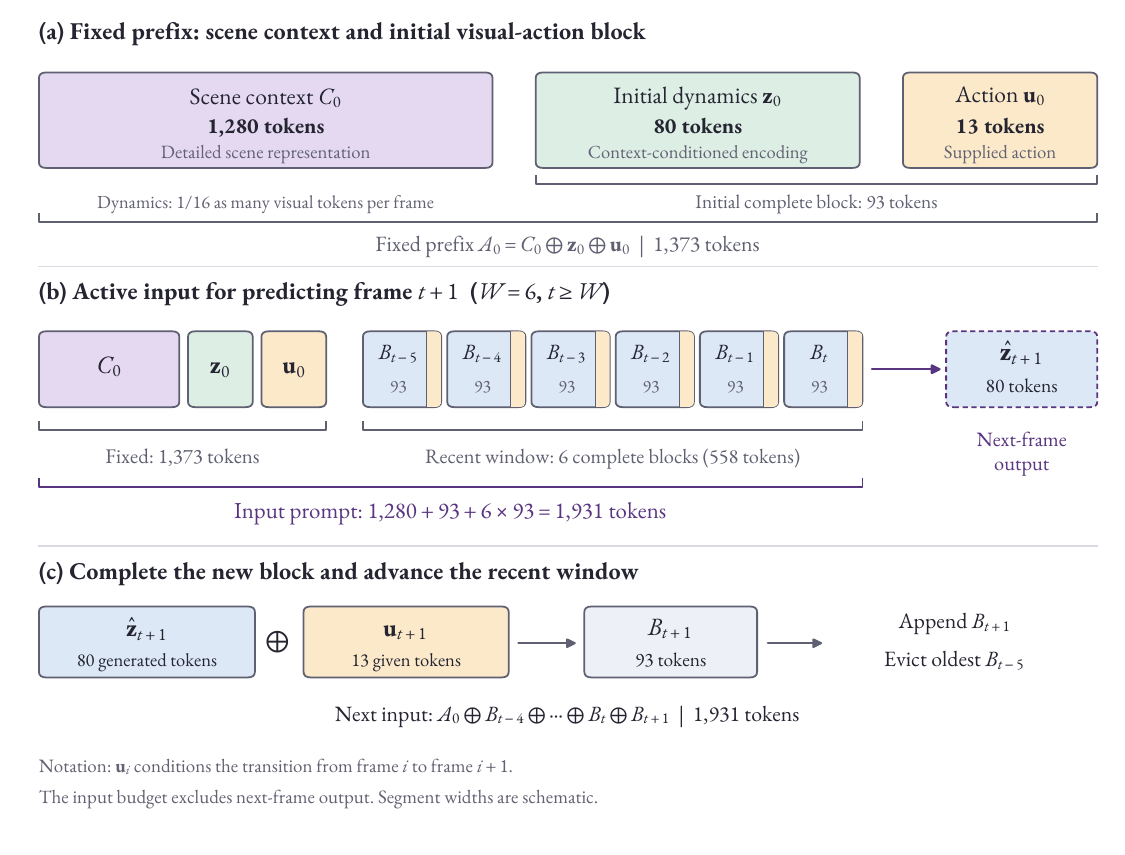}
  \caption{\textbf{Token organization for the 1,931-token ReCAP configuration.}
  (a) The fixed prefix comprises 1,280 scene-context tokens, 80 initial
  dynamics tokens, and 13 supplied action tokens, totaling 1,373 tokens.
  Each context-conditioned dynamics representation uses one-sixteenth
  as many visual tokens as the context representation.
  (b) Six recent complete visual--action blocks
  contribute 558 tokens, giving a 1,931-token input before next-frame
  generation. The next 80 visual tokens are outputs and are excluded from
  this input budget. (c) Before a further prediction, the generated visual
  tokens are paired with the next supplied action to complete a 93-token
  block, and the oldest recent block is evicted. Here, action $\mathbf{u}_i$
  conditions the transition from frame $i$ to frame $i+1$. Segment widths
  are schematic.}
  \label{fig:recap_token_sequence}
\end{figure}

%% file: Appendix/appendix_calvin_reproducibility.tex
\subsection{CALVIN Adaptation and Training Details}
\label{app:calvin-reproducibility}

\paragraph{Dataset and preprocessing.}
We use the official CALVIN D-to-D split (\texttt{task\_D\_D})~\citep{mees2022calvin}:
31 training rollouts (512,077 time steps) and four validation rollouts
(99,022). Each rollout is partitioned into non-overlapping clips of at most
128 frames, discarding remainders shorter than eight frames. This yields
4,013 training clips (512,072 frames) and 776 validation clips (99,022 frames).
Training samples contiguous eight-frame segments at stride one, without
crossing rollout boundaries.

Static-camera images ($200\times200$) are resized to $256\times256$, padded
by replicating 32 pixels at each horizontal edge to obtain $256\times320$,
and scaled to $[0,1]$. No image augmentation is applied.

\paragraph{Action conversion.}
The seven-dimensional normalized relative action is mapped to 13 slots:
translation in slots 4--6, rotation in 10--12, gripper in 3, and zeros in
0--2 and 7--9 (zero-based indexing). Each dimension uses 256 uniform bins,
with ranges estimated from training actions only and expanded by a 2\%
margin: $[-1.04,1.04]$ for active dimensions and $[-1,1]$ for unused slots.
Each visual--action block thus contains 80 dynamics tokens and 13 action tokens.

\paragraph{Stage 1: visual-tokenizer adaptation.}
We adapt the released RLVR-World RT-1 compressive FSQ tokenizer~\citep{wu2025rlvr}.
It uses a $32\times40$ context grid (1,280 tokens), an $8\times10$ dynamics
grid (80 tokens), and 4,375 codes per stream. Eight-frame segments contain
one context frame and seven dynamics frames. Table~\ref{tab:calvin-training-hparams}
gives optimization settings; AdamW uses $(\beta_1,\beta_2)=(0.9,0.999)$ and
$\epsilon=10^{-8}$, with seed 20260903. The objective combines unit-weight L1
and LPIPS terms with context/dynamics FSQ commitment losses. Context and
dynamics contributions are weighted by temporal occupancy ($1/8$ and $7/8$).
Adversarial training is disabled.

\begin{figure*}[t]
  \centering
  \includegraphics[width=\textwidth]{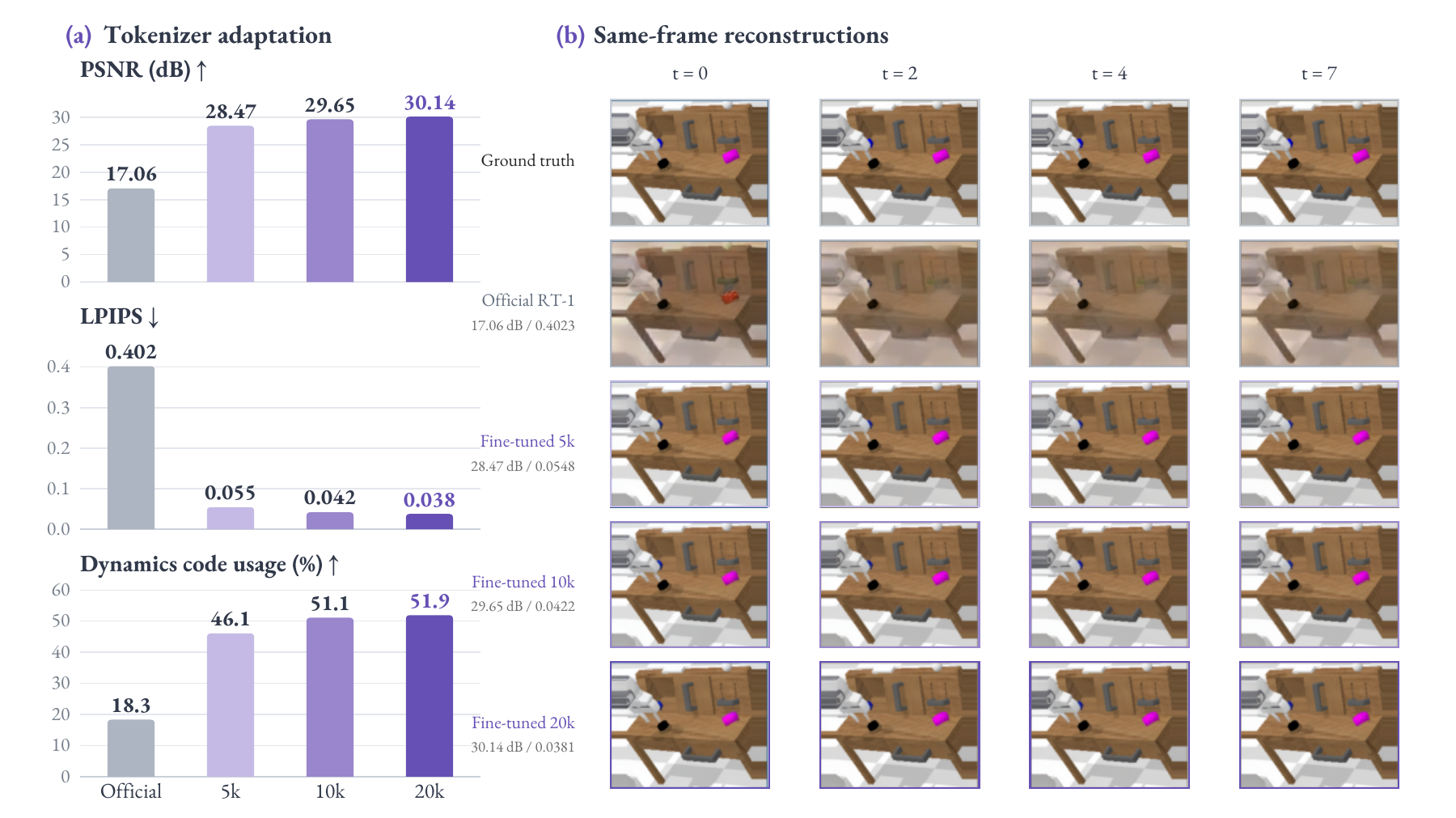}
  \caption{CALVIN tokenizer adaptation. (a) Reconstruction PSNR, LPIPS, and dynamics-code utilization measured on the same fixed set of 32 validation clips (256 frames) for the released RT-1 tokenizer and the 5k, 10k, and 20k checkpoints. (b) Same-frame reconstructions at $t\in\{0,2,4,7\}$. The adapted tokenizer progressively removes the color shift and object disappearance observed with the unadapted RT-1 tokenizer.}
  \label{fig:calvin-tokenizer-adaptation}
\end{figure*}

We evaluate checkpoints on 32 fixed, non-overlapping validation clips (eight
per rollout; 256 frames). The selected 20k tokenizer improves PSNR, SSIM,
and LPIPS from 17.064 dB, 0.7520, and 0.4023 to 30.138 dB, 0.9457, and 0.0381;
dynamics-code utilization rises from 18.3\% to 51.9\%
(Figure~\ref{fig:calvin-tokenizer-adaptation}).

\paragraph{Stage 2: action-conditioned world-model adaptation.}
We freeze the 20k tokenizer and adapt the released RT-1 world model using
causal next-token cross-entropy. Each eight-frame training sequence contains
1,280 context tokens and seven 93-token visual--action blocks (1,931 tokens).
Optimization settings are listed in Table~\ref{tab:calvin-training-hparams};
AdamW coefficients match Stage 1. Online validation uses two mini-batches
per evaluation.

\begin{figure*}[t]
  \centering
  \includegraphics[width=\textwidth]{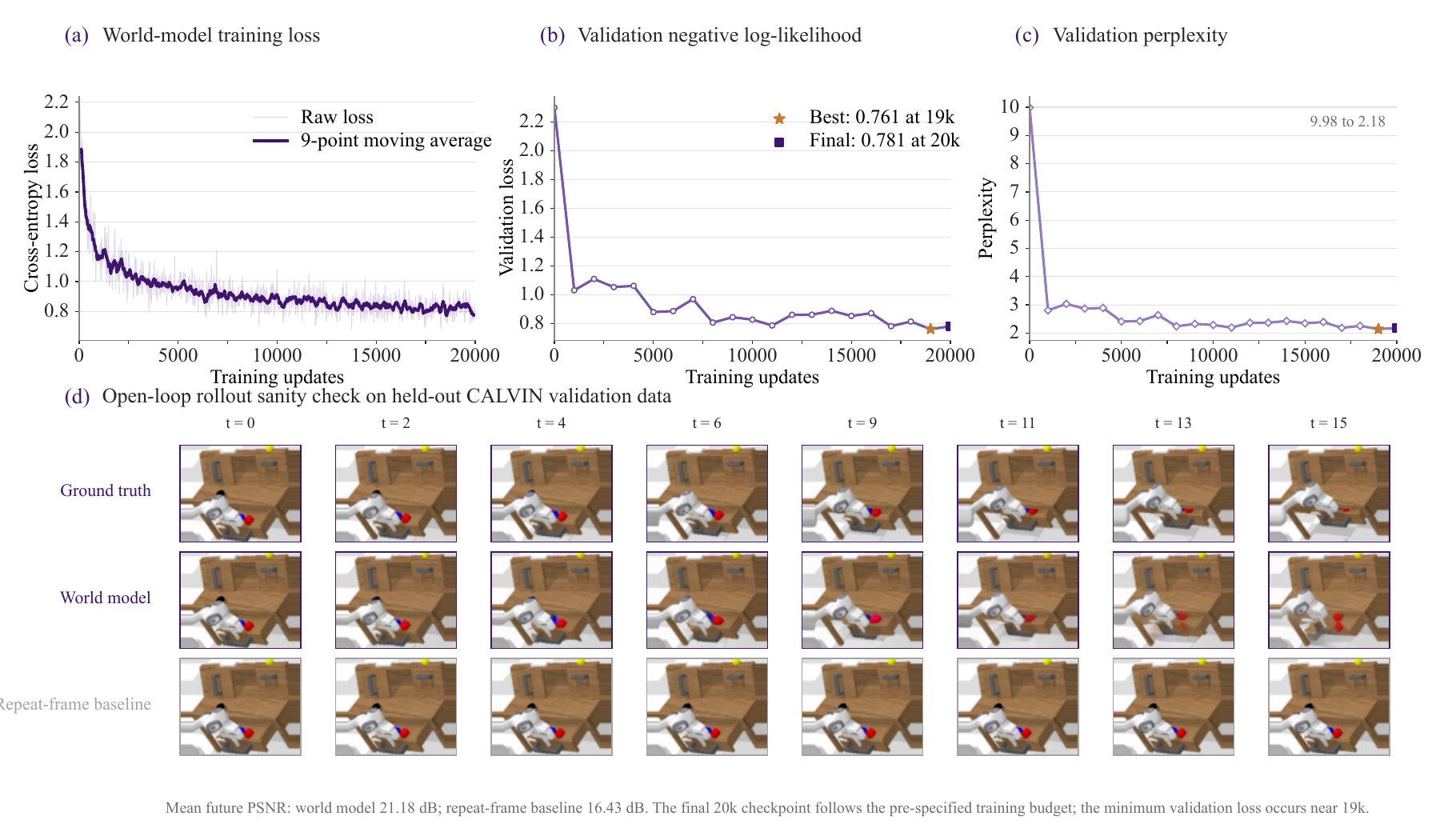}
  \caption{CALVIN world-model adaptation. (a) Raw training loss and its nine-point moving average. (b) Validation negative log-likelihood evaluated every 1,000 updates; the star marks the minimum near 19k and the square marks the pre-specified final 20k checkpoint. (c) Validation perplexity. (d) A 16-frame action-conditioned open-loop rollout on held-out CALVIN validation data. The model receives the first frame and ground-truth actions, then autoregressively feeds back predicted visual tokens. It achieves 21.18 dB mean future PSNR, compared with 16.43 dB for repeating the first frame.}
  \label{fig:calvin-world-model-training}
\end{figure*}

All CALVIN rollouts use the final 20k checkpoint, following the specified
training budget. Monitored validation loss decreases from 2.3010 to 0.7809
(minimum 0.7611 near 19k), and perplexity from 9.9837 to 2.1835.
Figure~\ref{fig:calvin-world-model-training} shows the training curves and a
held-out open-loop rollout.

\begin{table*}[t]
  \centering
  \small
  \caption{CALVIN adaptation and training hyperparameters. Batch size denotes the global batch across eight H20 GPUs.}
  \label{tab:calvin-training-hparams}
  \begin{tabular}{lcc}
    \toprule
    Setting & Tokenizer adaptation & World-model adaptation \\
    \midrule
    Initialization & RLVR-World RT-1 tokenizer & RLVR-World RT-1 world model \\
    Trainable component & Compressive tokenizer & Autoregressive transformer \\
    Training updates & 20,000 & 20,000 \\
    Segment / context frames & 8 / 1 & 8 / 1 \\
    Global batch size & 8 & 8 \\
    Optimizer & AdamW & AdamW \\
    Peak learning rate & $5\times10^{-6}$ & $1\times10^{-5}$ \\
    LR schedule & Cosine & Constant with warmup \\
    Warmup updates & 1,000 & 5,000 \\
    Weight decay & 0 & 0 \\
    Gradient clipping & 1.0 & 1.0 \\
    Precision & bfloat16 & bfloat16 \\
    Image augmentation & None & None \\
    Checkpoint / validation interval & 1,000 & 1,000 \\
    GPUs & 8$\times$ NVIDIA H20 & 8$\times$ NVIDIA H20 \\
    Wall-clock time & 2 h 08 min & 33 min 13 s \\
    \bottomrule
  \end{tabular}
\end{table*}

%% file: Appendix/appendix_libero_reproducibility.tex
\subsection{LIBERO-90 Adaptation and Training Details}
\label{app:libero-reproducibility}

\paragraph{Dataset and task-level split.}
LIBERO-90~\citep{liu2024libero} contains 90 tasks with 50 demonstrations each.
We lexicographically sort the task files and hold out nine evenly spaced
tasks, yielding 81 training tasks (4,050 episodes; 596,637 time steps) and
nine validation tasks (450 episodes; 72,406 time steps). Demonstrations
remain intact; training samples contiguous eight-frame segments at stride one.

We use the static \texttt{agentview\_rgb} camera. Native $128\times128$ images
are bilinearly resized to $256\times256$, edge-padded to $256\times320$ as in
CALVIN, and scaled to $[0,1]$, without augmentation.

\paragraph{Action conversion.}
We use the CALVIN seven-to-13-slot mapping and 256 uniform bins per dimension.
Ranges are fitted on training actions only, with a 2\% margin:
$[-1.04,1.04]$ for the gripper, $[-0.975,0.975]$ for translation,
$[-0.35394,0.38929]$ for roll, $[-0.39,0.39]$ for pitch and yaw, and $[-1,1]$
for unused slots.

\begin{figure*}[t]
  \centering
  \includegraphics[width=\textwidth]{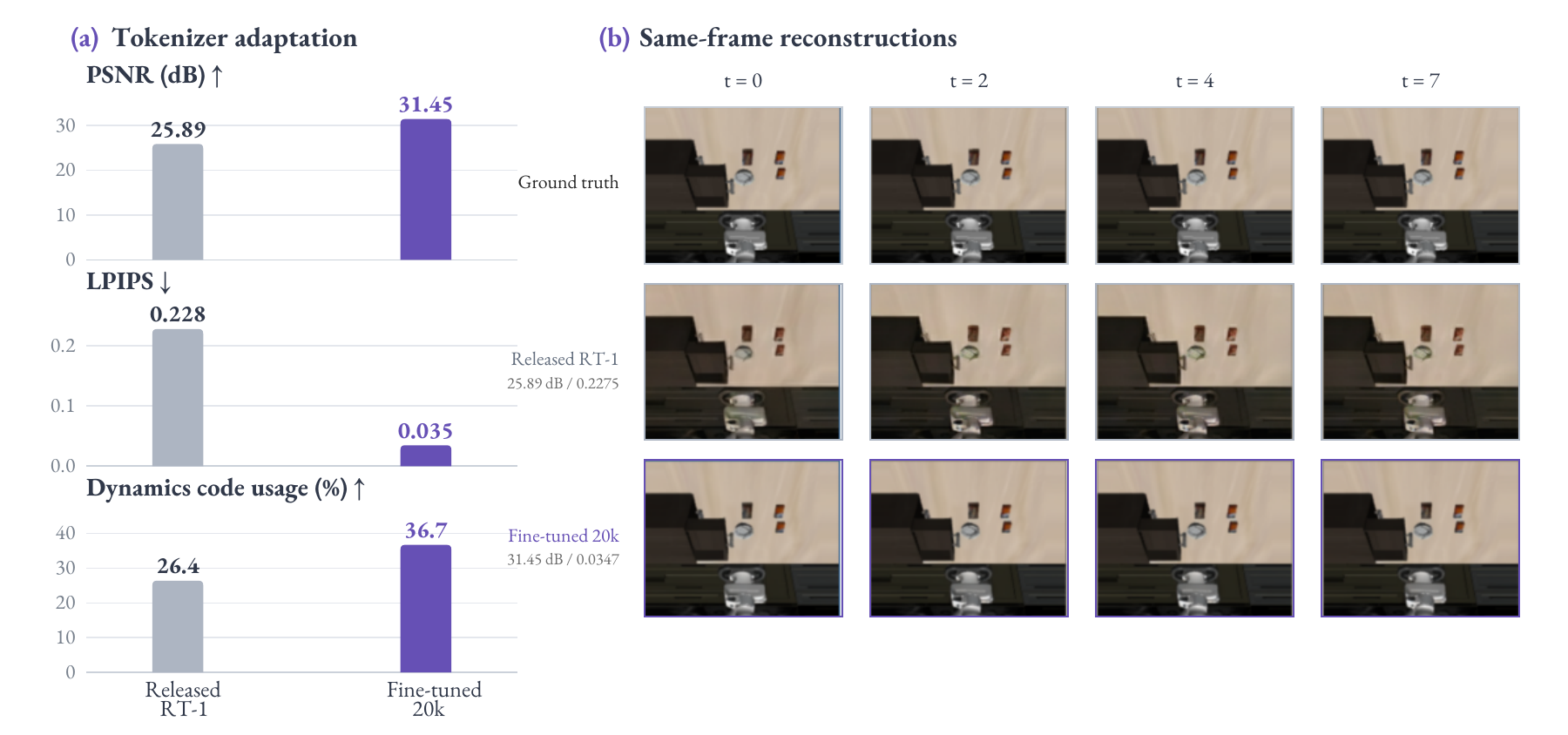}
  \caption{LIBERO-90 tokenizer adaptation. (a) Reconstruction PSNR, LPIPS, and dynamics-code utilization for the released RT-1 tokenizer and the final 20k LIBERO tokenizer, evaluated on the identical fixed set of 32 held-out clips (256 frames). (b) Same-frame reconstructions at $t\in\{0,2,4,7\}$. Adaptation substantially reduces color shift and preserves robot and object structure.}
  \label{fig:libero-tokenizer-adaptation}
\end{figure*}

\paragraph{Stage 1: visual-tokenizer adaptation.}
Tokenizer initialization, representation, loss, optimizer coefficients,
and seed match CALVIN (Appendix~\ref{app:calvin-reproducibility});
Table~\ref{tab:libero-training-hparams} lists the training settings.
We compare the released tokenizer and the final 20k checkpoint on 32 fixed,
non-overlapping clips from eight held-out episodes (256 frames). PSNR,
SSIM, and LPIPS improve from 25.8858 dB, 0.9062, and 0.2275 to 31.4548 dB,
0.9669, and 0.0347, while dynamics-code utilization rises from 26.35\% to
36.66\% (Figure~\ref{fig:libero-tokenizer-adaptation}). The final checkpoint
follows the specified training budget; intermediate images from random
validation batches are not used for checkpoint selection.

\paragraph{Stage 2: action-conditioned world-model adaptation.}
We freeze the 20k tokenizer and adapt the RT-1 world model using the same
1,931-token sequences, causal objective, and optimization recipe as CALVIN
(Table~\ref{tab:libero-training-hparams}), with seed 20260903.
Online validation uses two mini-batches per evaluation. Monitored loss
decreases from 1.7984 to 0.8089 and perplexity from 6.0401 to 2.2453;
the minimum loss is 0.7482 near 17k (perplexity 2.1133). All LIBERO experiments
use the final 20k checkpoint specified by the training budget.

\begin{figure*}[t]
  \centering
  \includegraphics[width=\textwidth]{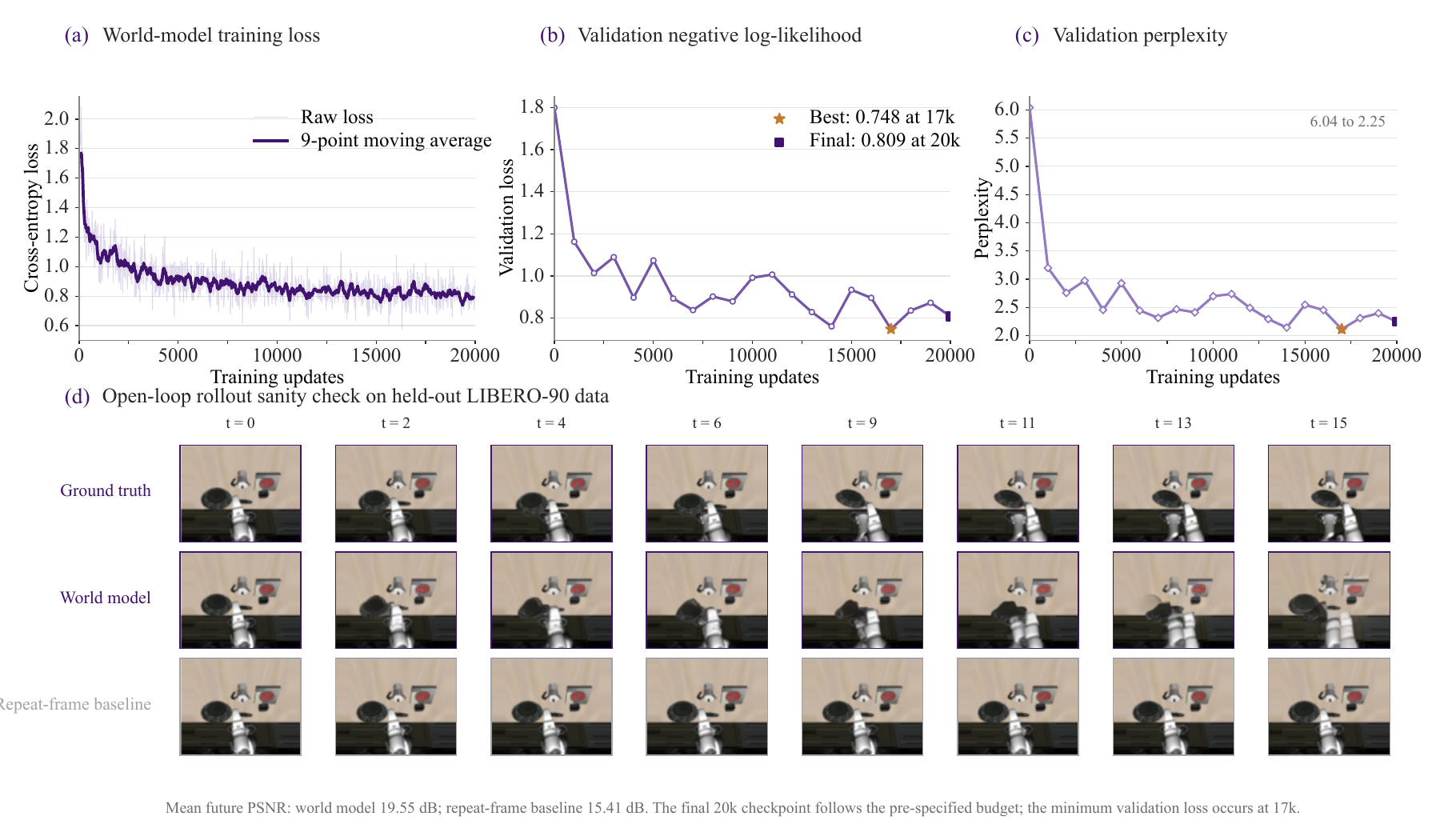}
  \caption{LIBERO-90 world-model adaptation. (a) Raw training loss and its nine-point moving average. (b) Validation negative log-likelihood evaluated every 1,000 updates; the star marks the minimum near 17k and the square marks the pre-specified final 20k checkpoint. (c) Validation perplexity. (d) A 16-frame action-conditioned visual open-loop rollout on a held-out high-motion LIBERO-90 episode. The model observes the first frame, receives ground-truth future actions, and autoregressively feeds back its visual predictions.}
  \label{fig:libero-world-model-training}
\end{figure*}

On a held-out high-motion episode, the model receives one observed frame
and recorded future actions. Mean future PSNR is 19.549 dB for a 16-frame
open-loop rollout, versus 15.411 dB for repeating the observed frame;
at 32 frames, the values are 15.193 and 13.719 dB.
Figure~\ref{fig:libero-world-model-training} shows the training curves and
16-frame rollout.

\begin{table*}[t]
  \centering
  \small
  \caption{LIBERO-90 adaptation and training hyperparameters. Batch size denotes the global batch across eight H20 GPUs.}
  \label{tab:libero-training-hparams}
  \begin{tabular}{lcc}
    \toprule
    Setting & Tokenizer adaptation & World-model adaptation \\
    \midrule
    Initialization & RLVR-World RT-1 tokenizer & RLVR-World RT-1 world model \\
    Trainable component & Compressive tokenizer & Autoregressive transformer \\
    Training updates & 20,000 & 20,000 \\
    Segment / context frames & 8 / 1 & 8 / 1 \\
    Global batch size & 8 & 8 \\
    Optimizer & AdamW & AdamW \\
    Peak learning rate & $5\times10^{-6}$ & $1\times10^{-5}$ \\
    LR schedule & Cosine & Constant with warmup \\
    Warmup updates & 1,000 & 5,000 \\
    Weight decay & 0 & 0 \\
    Gradient clipping & 1.0 & 1.0 \\
    Precision & bfloat16 & bfloat16 \\
    Image augmentation & None & None \\
    Checkpoint / validation interval & 1,000 & 1,000 \\
    GPUs & 8$\times$ NVIDIA H20 & 8$\times$ NVIDIA H20 \\
    Wall-clock time & 2 h 07 min 55 s & 32 min 33 s \\
    \bottomrule
  \end{tabular}
\end{table*}

%% file: Appendix/appendix_bridge_reproducibility.tex
\subsection{BridgeV2 Adaptation and Short-Horizon Evaluation}
\label{app:bridge-reproducibility}

\paragraph{Data and action representation.}
We preserve the train/test split of BridgeV2 TFDS 0.1.0~\citep{walke2023bridgedata}:
25,460 training episodes (864,292 time steps) and 3,475 validation episodes
(118,603 time steps), with mean length approximately 34 frames. Training
samples contiguous eight-frame segments at stride one within episodes.
Images use the native $256\times320$ geometry and the same 1,280-token
context, 80-token dynamics representation, and 4,375-code FSQ streams as CALVIN.

Bridge's action mapping differs from CALVIN and LIBERO: translation occupies
slots 0--2, rotation 3--5, gripper 6, zeros 7--11, and termination 12.
Each dimension uses 256 bins with ranges fitted on training actions only,
preserving the 93-token visual--action block.

For the additional qualitative study, we take the first 24 validation files
in lexicographic order with at least 33 frames and retain their first 33
frames: one observation and 32 prediction targets. This candidate pool is
separate from the main-text BridgeV2 evaluation sets.

\paragraph{Stage 1: tokenizer adaptation.}
We fine-tune the released RLVR-World RT-1 tokenizer~\citep{wu2025rlvr} with an
adversarial reconstruction objective. Table~\ref{tab:bridge-training-hparams}
lists the optimization settings. Both AdamW optimizers use
$(\beta_1,\beta_2)=(0.9,0.999)$ and $\epsilon=10^{-8}$; gradient checkpointing
is enabled and the seed is 20260906. The final checkpoint is shared step
60,000, reached in approximately 18 h 49 min.

Before shared step 10,000, only the tokenizer is updated. Thereafter,
generator and discriminator branches alternate after groups of four
accumulation micro-batches. Each learning-rate schedule advances only with
its corresponding optimizer. Thus, the 60k shared counter does not denote
60,000 generator updates.

\paragraph{Augmentation and reconstruction objective.}
Tokenizer training uses clip-consistent color jitter: brightness, contrast, and saturation factors in $[0.9,1.1]$, and hue shifts in $[-0.05,0.05]$. Random resized crops use area scale $[0.8,1.0]$ and aspect ratio $[1.0,1.4]$, with output size $256\times320$. The same sampled transform is applied across each clip. The online validation loader inherits these augmentation settings; the fixed-subset reconstruction evaluation below instead uses unchanged input frames.

Let $x_c$ denote the context frame and $x_{1:7}$ the dynamics frames, with reconstructions $\hat x$. For this run, the generator objective is
\begin{equation}
\begin{split}
\mathcal L_G ={}& \mathcal L_1(x_c,\hat x_c)
 + \mathcal L_1(x_{1:7},\hat x_{1:7})\\
 &+ \mathcal L_{\mathrm{LPIPS}}(x_c,\hat x_c)
 + \mathcal L_{\mathrm{LPIPS}}(x_{1:7},\hat x_{1:7})\\
 &+0.1\,\mathbb{1}[s\geq10{,}000]\,
 \mathcal L_{\mathrm{adv}},\qquad
 \mathcal L_{\mathrm{adv}}=-\mathbb E[D(\hat x)].
\end{split}
\label{eq:bridge-tokenizer-objective}
\end{equation}
Each dynamics term averages over seven frames. Context and dynamics terms
have unit weight, without temporal-occupancy balancing. FSQ commitment
losses are zero, and the adversarial weight is fixed at 0.1.

\paragraph{Adversarial objective.}
Following adversarial perceptual tokenization~\citep{esser2021taming}, a
frame-level discriminator processes real and reconstructed context and
dynamics images. It uses six stride-two convolutional stages, spectral
normalization, LeakyReLU activations, intermediate instance normalization,
and a spatial logit head. With reconstructed frames detached, its loss is
\begin{equation}
\mathcal L_D=\mathbb E[(1-D(x))_+ +(1+D(\hat x))_+]
+\mathcal R_{\mathrm{GP}}.
\end{equation}
The real-input gradient penalty has coefficient 10 and unit target norm,
using gradients of summed logits with respect to real dynamics inputs.
The discriminator is used only during tokenizer adaptation. Full Context
and \method{} share the resulting frozen tokenizer and transformer;
we do not separately estimate the effect of adversarial training.

\paragraph{Tokenizer evaluation.}
The final 60k tokenizer is evaluated on 32 eight-frame clips from the first
eight validation episodes (four fixed starts per episode; some clips overlap).
Across 256 frame evaluations, PSNR, SSIM, and LPIPS are 27.0667 dB, 0.8336,
and 0.1089; dynamics-only values are 26.7168 dB, 0.8286, and 0.1115.
Dynamics-code utilization is 65.19\%. These measure reconstruction from real
frames, not open-loop prediction.

\begin{figure*}[htbp]
\centering
\includegraphics[width=\textwidth]{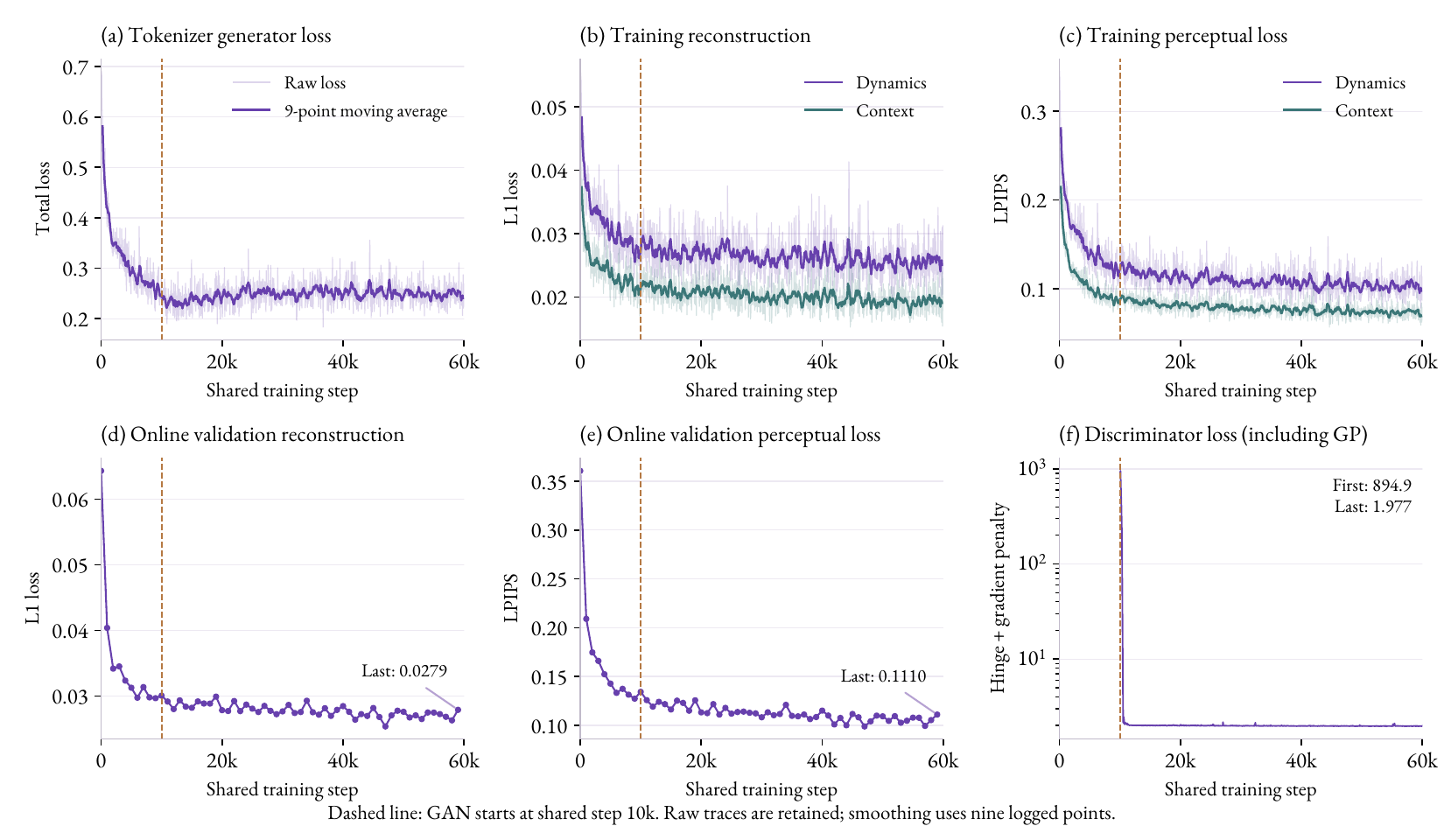}
\caption{\textbf{BridgeV2 tokenizer training.} (a) Total generator loss; (b--c) context and dynamics reconstruction and perceptual losses; (d--e) online validation reconstruction and perceptual losses; (f) discriminator hinge loss plus gradient penalty (logarithmic vertical axis). Dashed lines mark adversarial training activation at shared step 10k. Faint training traces show raw logged values; solid training traces use a nine-point moving average. Validation points are unsmoothed and end at step 59,001. The shared counter includes alternating generator/discriminator updates after GAN activation.}
\label{fig:bridge-tokenizer-training}
\end{figure*}

Figure~\ref{fig:bridge-tokenizer-training} shows the optimization curves.
Online validation reconstruction loss decreases from 0.0643 to 0.0279 and
LPIPS from 0.3608 to 0.1110 at the last evaluation (step 59,001). These
augmented-batch evaluations differ from the unchanged fixed subset above.
The generator objective also changes at adversarial activation (step 10k).

\FloatBarrier

\paragraph{Stage 2: action-conditioned world-model adaptation.}
We freeze the 60k tokenizer and adapt the released RT-1 transformer
(127,101,696 parameters) for 100,000 updates using causal next-token
cross-entropy on 1,931-token, eight-frame sequences.
Table~\ref{tab:bridge-training-hparams} gives training settings; AdamW
coefficients match Stage 1, with seed 20260906 and no image augmentation
or adversarial loss. Online validation uses two mini-batches per evaluation,
so its fluctuations do not estimate full-validation-set performance.

\begin{figure*}[htbp]
\centering
\includegraphics[width=\textwidth]{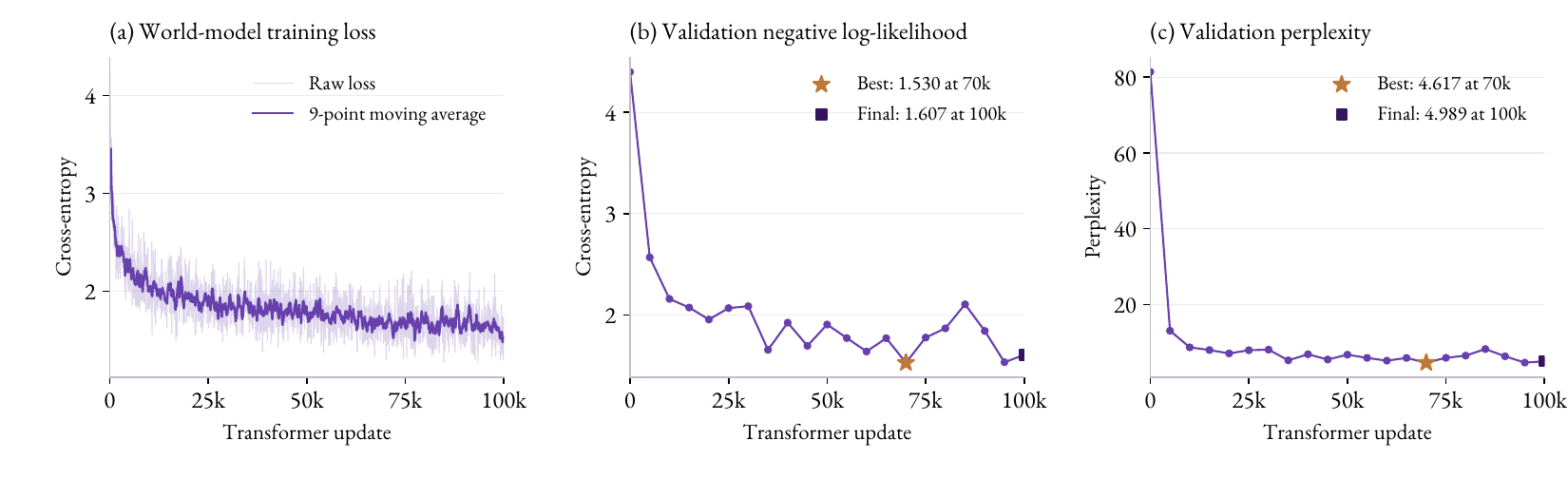}
\caption{\textbf{BridgeV2 world-model training.} (a) Raw causal cross-entropy and its nine-point moving average over 100k transformer updates. (b) Validation negative log-likelihood and (c) validation perplexity. Stars indicate the lowest monitored validation loss near 70k; squares mark the final 100k checkpoint used for the rollout comparisons. All validation points are plotted without smoothing, including the final evaluation at 100,000.}
\label{fig:bridge-world-training}
\end{figure*}

Training loss decreases from 4.2416 to 1.5072; monitored validation loss
from 4.3995 to 1.6072 and perplexity from 81.4104 to 4.9888
(Figure~\ref{fig:bridge-world-training}). Although validation loss reaches
its minimum of 1.5297 near 70k (perplexity 4.6168), rollout comparisons use
the final 100k checkpoint specified by the training budget.

\FloatBarrier

\begin{table}[ht]
\centering
\small
\caption{BridgeV2 adaptation hyperparameters. Tokenizer steps use the shared generator/discriminator counter; world-model steps are transformer updates.}
\label{tab:bridge-training-hparams}
\begin{tabular}{lcc}
\toprule
Setting & Tokenizer adaptation & World-model adaptation \\
\midrule
Initialization & RT-1 tokenizer & RT-1 world model \\
Checkpoint counter & 60,000 (shared) & 100,000 \\
Segment / context frames & 8 / 1 & 8 / 1 \\
Per-device batch / accumulation & 1 / 4 & 1 / 1 \\
Global batch per optimizer update & 32 & 8 \\
Optimizer / weight decay & AdamW / 0 & AdamW / 0 \\
Peak learning rate & $5\times10^{-5}$ (G and D) & $10^{-5}$ \\
LR schedule & Cosine (G), constant (D) & Constant \\
Configured warmup & 5,000 per schedule & 5,000 \\
Gradient clipping & 1.0 & 1.0 \\
Precision & BF16; TF32 enabled & BF16 \\
Adversarial start / weight & 10,000 / 0.1 & Disabled \\
Discriminator depth / GP weight & 6 / 10 & -- \\
Image augmentation & Color jitter and crop & None \\
Checkpoint / validation interval & 2,000 / 1,000 & 5,000 / 5,000 \\
GPUs & 8 H20 & 8 H20 \\
\bottomrule
\end{tabular}
\end{table}

\FloatBarrier
\paragraph{Paired inference protocol.}
The model observes $\mathbf{o}_0$ and predicts $\mathbf{o}_{1:32}$ from recorded
actions. The initial observation supplies both context and the initial
dynamics anchor. Full Context and \method{} share the checkpoint, observation,
actions, and sampling settings (Table~\ref{tab:bridge-additional-settings}).
This reference implementation uses PyTorch/Transformers on one H20 in FP32,
with TF32 disabled. Full Context reuses its KV cache incrementally.
\method{} retains the fixed anchor and six recent complete blocks; after
eviction it rebuilds KV and rebases positions, without re-encoding pixels.
Its prompt is bounded at 1,931 tokens. The methods produce identical visual
tokens for the first seven predictions, before eviction. These timings are
separate from the BF16 vLLM efficiency benchmark in RQ4.

\paragraph{Metrics and qualitative selection.}
We average full-image PSNR, SSIM (11-pixel window, $\sigma=1.5$), and VGG-LPIPS
over the 32 predicted frames, excluding the observation. Figure~\ref{fig:appendix_strong_bridge}
shows the three candidates with the largest relative LPIPS reduction among
24 episodes. Metadata supplies display labels only. These selected-case
gains are not full-validation-set estimates; remaining duplication and
shape errors also limit conclusions about task success.

\begin{table}[ht]
\centering
\small
\caption{BridgeV2 settings for the additional $H=32$ comparisons.}
\label{tab:bridge-additional-settings}
\begin{tabular}{ll}
\toprule
Setting & Value \\
\midrule
Tokenizer / world-model checkpoint & 60k / 100k \\
World-model parameters & 127.1M \\
Training GPUs / global batch / segment & 8 H20 / 8 / 8 frames \\
World-model learning rate / precision & $10^{-5}$ / BF16 \\
Observed / predicted frames & 1 / 32 \\
Inference precision & FP32, TF32 disabled \\
Sampling seed / temperature / top-$k$ & 20260906 / 1.0 / 100 \\
Recent window / maximum prompt & 6 blocks / 1,931 tokens \\
Candidate pool / displayed examples & 24 / 3 \\
\bottomrule
\end{tabular}
\end{table}

%% file: Appendix/appendix_additional_qualitative.tex
\section{Additional Qualitative Rollouts}
\label{app:additional_qualitative}

\begin{figure*}[!htbp]
    \centering
    \includegraphics[width=\textwidth]{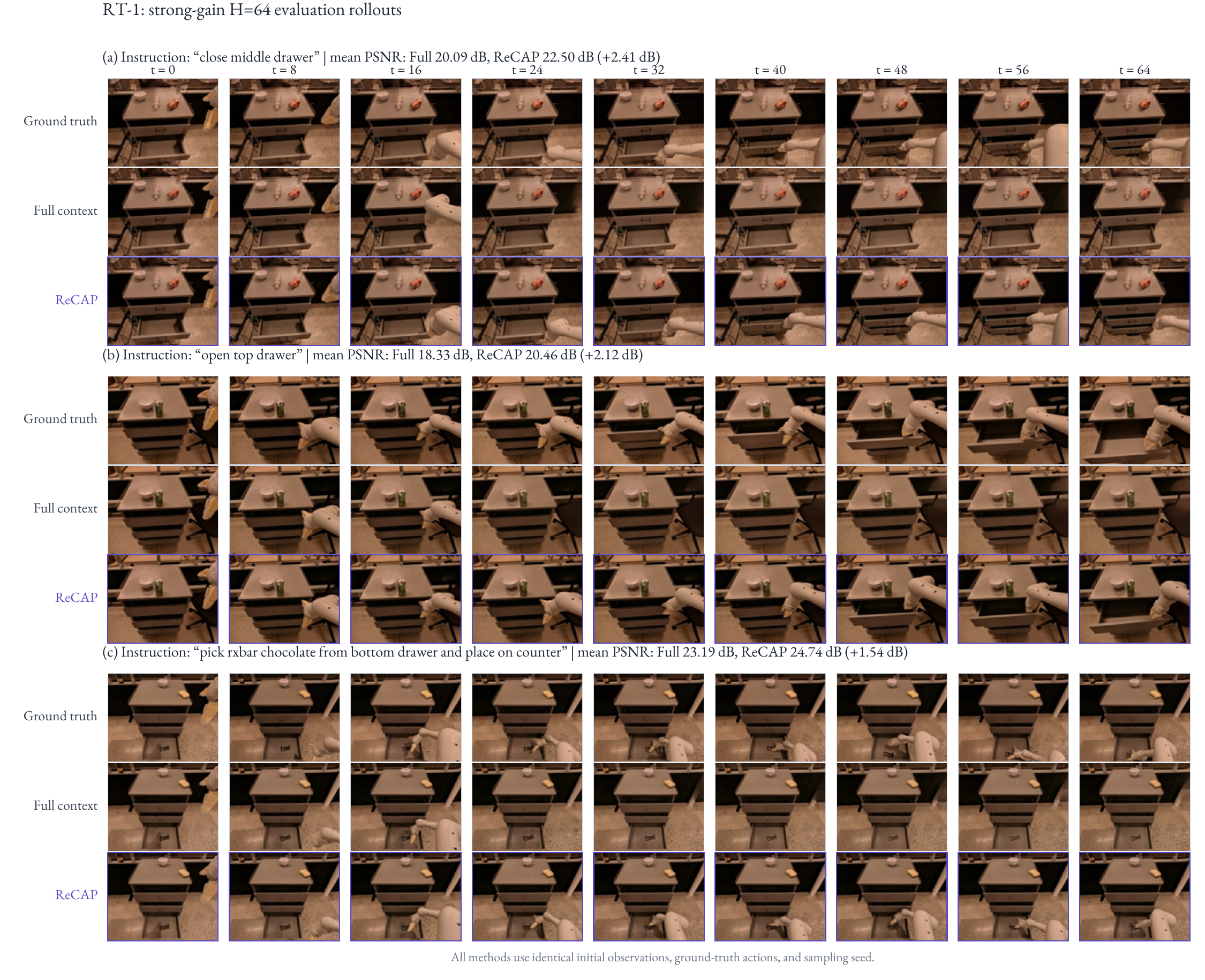}
    \caption{\textbf{\textit{Additional RT-1 qualitative comparisons.}}
    Nine time points are shown for each rollout.  Across these three
    instruction-conditioned examples, \method{} improves mean PSNR over Full
    Context by $+2.41$, $+2.12$, and $+1.54$\,dB, respectively.}
    \label{fig:appendix_strong_rt1}
\end{figure*}

\begin{figure*}[p]
    \centering
    \includegraphics[width=\textwidth]{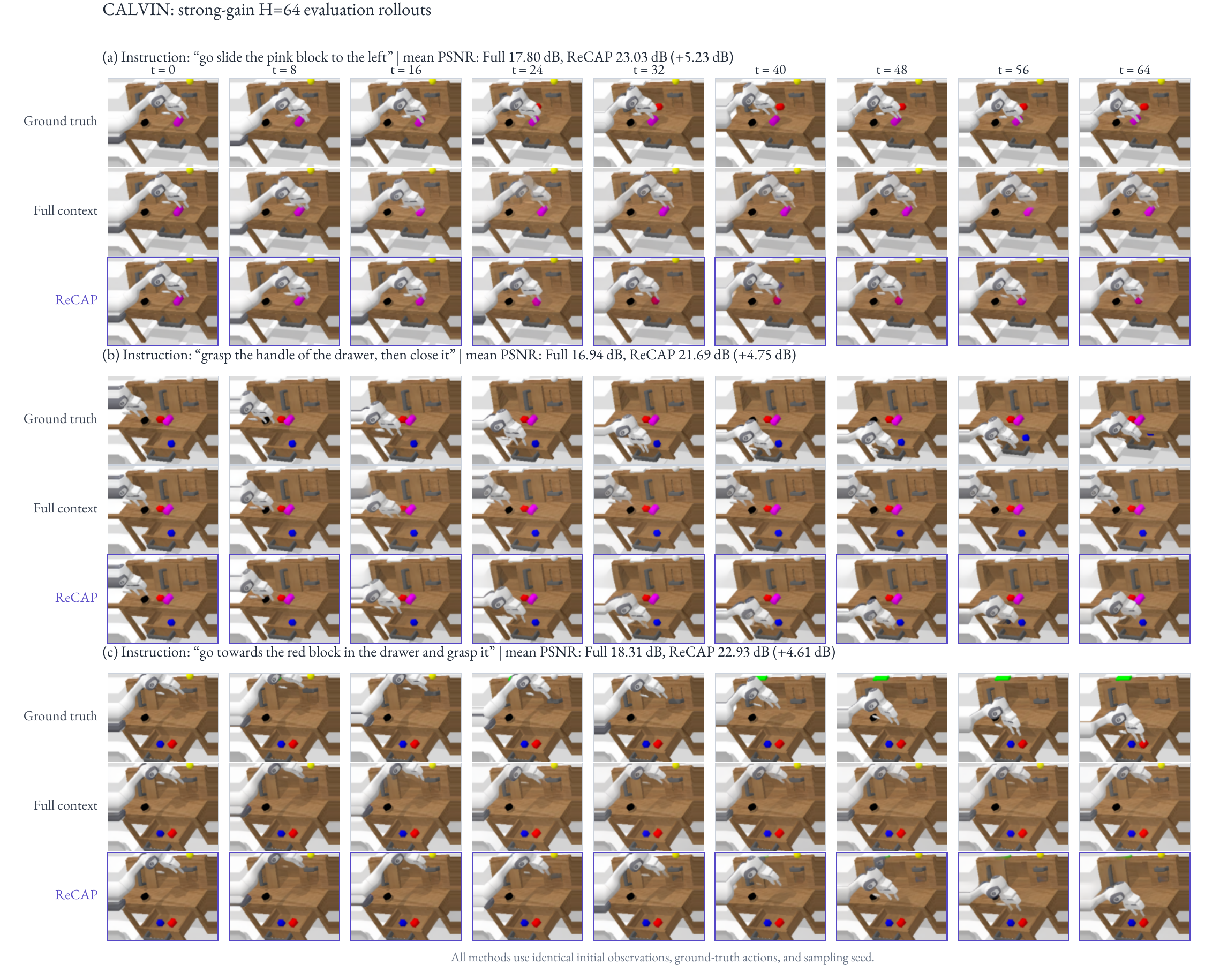}
    \caption{\textbf{\textit{Additional CALVIN validation comparisons.}}
    The examples are drawn from annotated validation trajectories.  \method{}
    improves mean PSNR over Full Context by $+5.23$, $+4.75$, and
    $+4.61$\,dB, respectively.}
    \label{fig:appendix_strong_calvin}
\end{figure*}

\begin{figure*}[p]
    \centering
    \includegraphics[width=\textwidth]{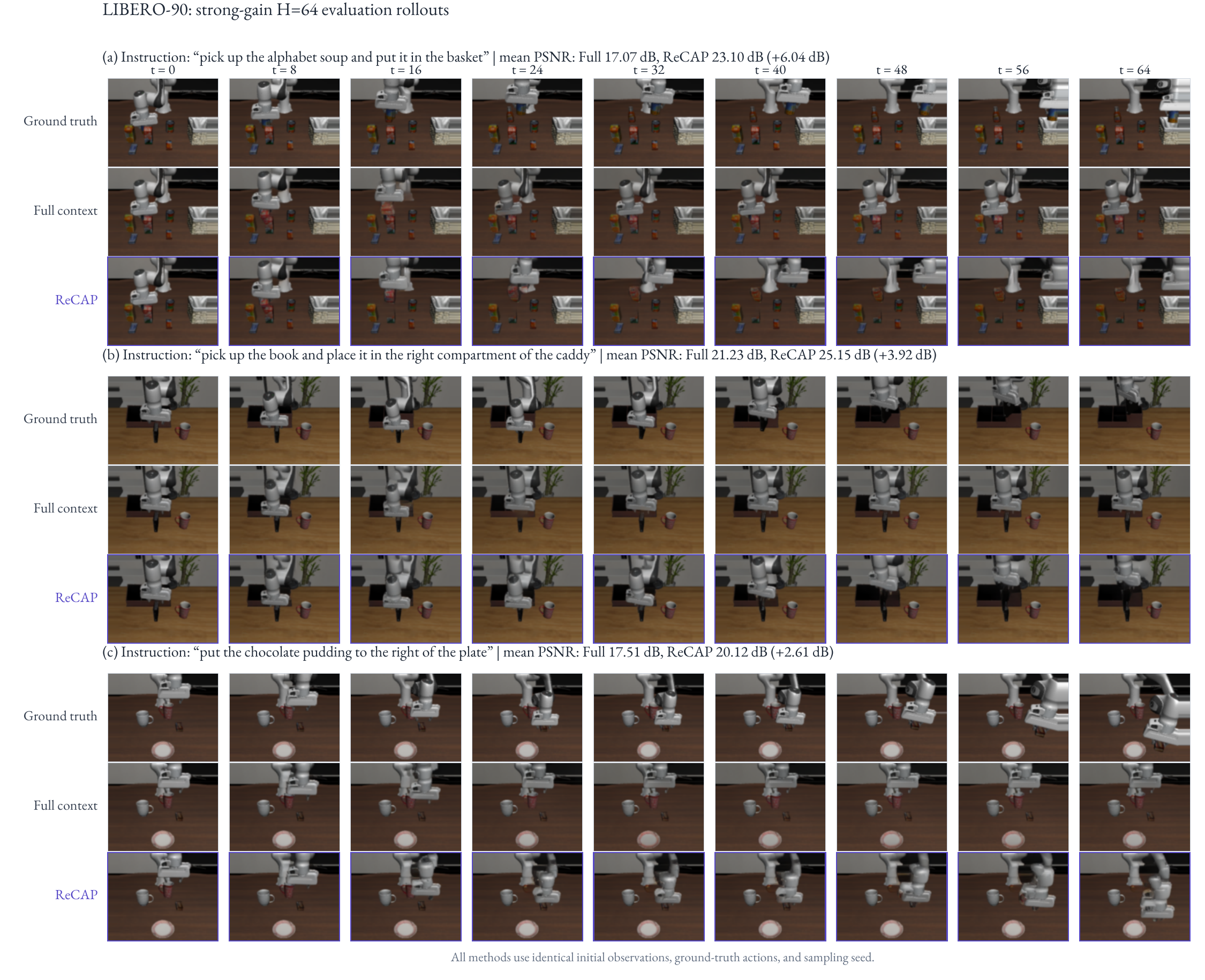}
    \caption{\textbf{\textit{Additional LIBERO-90 held-out-task comparisons.}}
    The three examples use distinct language instructions from the
    task-disjoint validation split.  \method{} improves mean PSNR over Full
    Context by $+6.04$, $+3.92$, and $+2.61$\,dB, respectively.  Frames are
    vertically flipped for canonical display orientation only.}
    \label{fig:appendix_strong_libero}
\end{figure*}

\begin{figure*}[p]
    \centering
    \includegraphics[width=\textwidth,height=0.82\textheight,keepaspectratio]{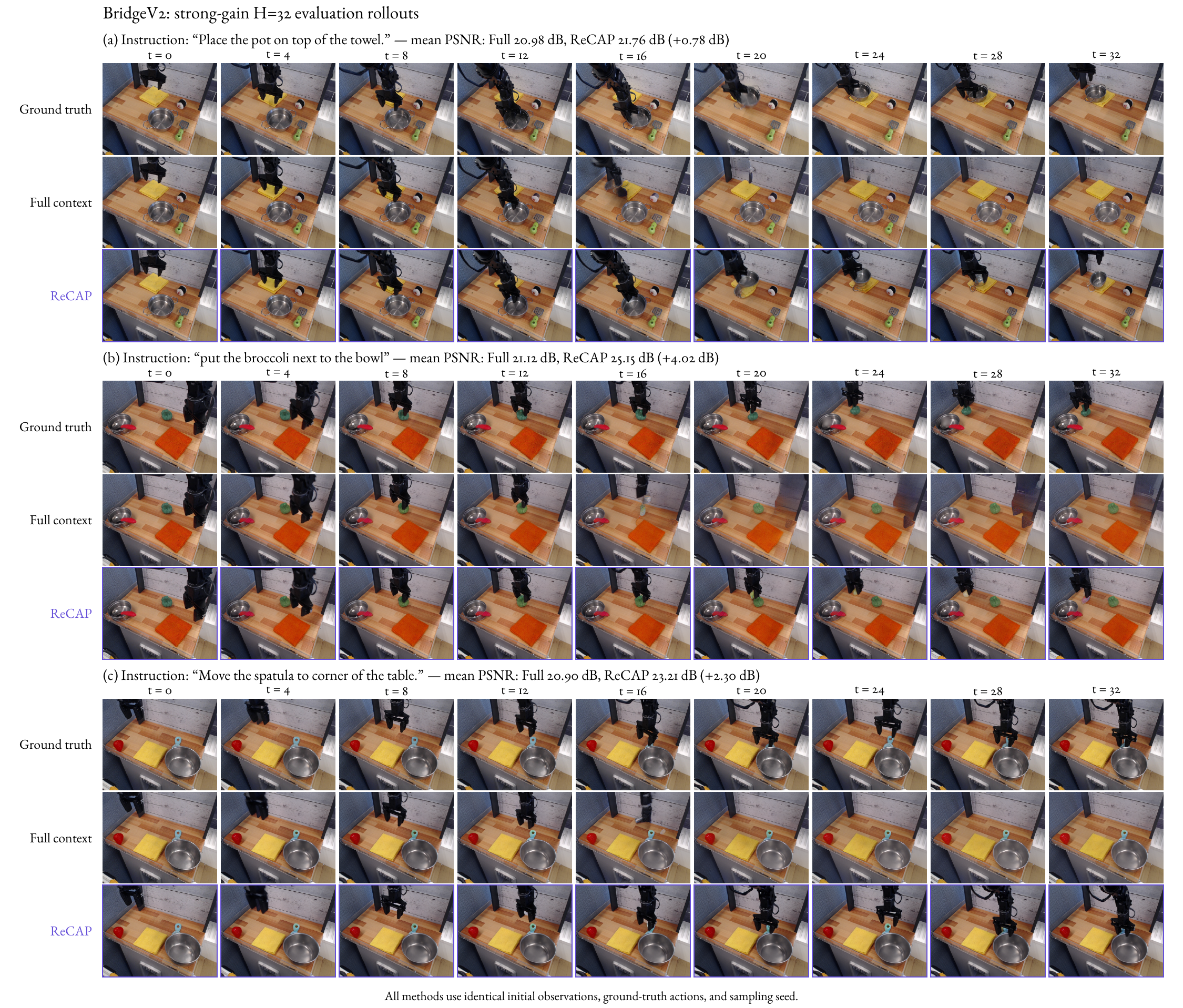}
    \caption{\textbf{\textit{Additional BridgeV2 validation comparisons.}}
    Three high-gain examples selected from a fixed pool of 24 candidates, with
    nine displayed time points ($t=0,4,8,12,16,20,24,28,32$) over 32 predicted frames. Rows show ground truth,
    Full Context, and \method{}. Mean future PSNR improves by $+0.78$, $+4.02$,
    and $+2.30$\,dB; mean LPIPS decreases by $26.7\%$, $32.5\%$, and $23.3\%$,
    respectively. Metrics use all 32 predictions, excluding $t=0$. Task text
    identifies the episodes and is not a model input.}
    \label{fig:appendix_strong_bridge}
\end{figure*}

Figures~\ref{fig:appendix_strong_rt1}--\ref{fig:appendix_strong_bridge}
compare matched rollouts with identical initial observations, recorded actions,
and sampling seeds: 64-frame rollouts for RT-1, CALVIN, and LIBERO-90, and
32-frame predictions for BridgeV2 (Appendix~\ref{app:bridge-reproducibility}).
Task labels come from dataset annotations. LIBERO images are vertically flipped
for display only; inference and metrics use the original arrays.

%% file: Appendix/appendix_vlm_protocol.tex
\section{VLM Evaluation Protocol}
\label{app:vlm_protocol}

\subsection{Evaluation Scope and Inputs}
\label{app:vlm_scope}

We evaluate 200 held-out RT-1 episodes, each with paired Full Context and
\method{} predictions sharing the initial observations, actions, and sampling
seed (400 videos per judge). Videos contain 26--113 frames (median 42) at
$256\times320$ resolution and are encoded as H.264 at 25 frames/s before sampling.

Each judge receives one video's sampled frames and the episode instruction,
without method identity, the competing prediction, or ground-truth video.
Scores are compared afterward across matched episodes. The instruction
specifies the evaluation goal; it is not a language input to the world model.
An additional 80-pair human comparison and sensitivity analysis appear in
Appendices~\ref{app:human_vlm_agreement} and~\ref{app:vlm_sensitivity}.

\subsection{Deterministic Frame Sampling}
\label{app:vlm_sampling}

Let $N$ be the decoded MP4 frame count. If $N\leq16$, all frames are used.
Otherwise, the evaluator selects the following zero-based indices in
temporal order:
\begin{equation}
    \mathcal{I}(N)=
    \left\{\left\lfloor\frac{iN}{16}\right\rfloor:
    i=0,\ldots,15\right\}.
    \label{eq:vlm_sampling}
\end{equation}
The same ordered RGB images at original resolution are shared across all
judges before model-specific preprocessing. Sampling is deterministic, with
no key-frame detection; the budget is 16 images, whose visual-token counts
vary by model. The final video frame is not necessarily included, so the
rubric's ``last frame'' refers to the last sampled image.

\subsection{Judge-Specific Preprocessing and Generation}
\label{app:vlm_backends}

\paragraph{Qwen3-VL-8B.}
We use the native chat template and smart resize, with sampling disabled
and at most 400 output tokens.

\paragraph{InternVL3-8B.}
Frames are bicubically resized to $448\times448$, normalized with ImageNet
means $(0.485,\allowbreak 0.456,\allowbreak 0.406)$ and standard deviations $(0.229,\allowbreak 0.224,\allowbreak 0.225)$,
and converted to BF16. Each frame uses one tile, without dynamic tiling,
and is labeled in temporal order. The square resize changes the source aspect ratio.

\paragraph{GPT-5.5.}
The API receives the same sampled images and rubric with provider-default
decoding parameters; repeated responses need not be identical.

\subsection{Shared Rubric and Required Output}
\label{app:vlm_rubric}

The fixed prompt requires an integer score from 1 to 10 and an evidence-based
reason for each of four criteria, returned as JSON. Judgments must use only
visible evidence. The episode \texttt{\{instruction\}} is the sole runtime
text substitution. Table~\ref{tab:vlm_rubric} summarizes the rubric;
Appendix~\ref{app:vlm_prompt} reproduces the complete prompt.

\begin{table}[ht]
    \centering
    \caption{\textbf{\textit{Shared VLM evaluation rubric.}}
    Score bands summarize the prompt; the mandatory score caps also apply. Higher values indicate better observed behavior.}
    \label{tab:vlm_rubric}
    \footnotesize
    \renewcommand{\arraystretch}{1.16}
    \setlength{\tabcolsep}{3pt}
    \begin{tabularx}{\textwidth}{@{}l*{4}{>{\raggedright\arraybackslash}X}@{}}
        \toprule
        \textbf{Score} & \textbf{Interaction quality}
        & \textbf{Instruction following} & \textbf{Task progress}
        & \textbf{Motion naturalness} \\
        \midrule
        1--2 & Penetration or absent proper contact
        & Wrong action, object, or scene & No visible progress
        & Severe jumps or jitter; or a frozen arm \\
        3--4 & Contact with implausible physical response
        & Related behavior with major errors & Motion without meaningful progress
        & Frequent irregular motion; or tiny, aimless movements \\
        5--6 & Mostly plausible, with minor interaction errors
        & General intent followed with execution errors & Partial completion
        & Meaningful motion with occasional stutter or kinematic issues \\
        7--8 & Realistic contact and object response
        & Mostly correct, with minor deviations & Mostly completed
        & Meaningful, smooth motion with minor imperfections \\
        9--10 & Visually indistinguishable from real interaction
        & All instructed elements correctly executed & Fully completed
        & Meaningful, smooth, physically plausible motion \\
        \bottomrule
    \end{tabularx}
\end{table}

\paragraph{Mandatory caps and score guidance.}
The \emph{hallucination check} caps Instruction Following at 4 when human
hands replace the expected robotic arm. The \emph{static penalty} caps
Motion Naturalness at 4 when the arm shows essentially no meaningful
motion or interaction, even if the scene appears smooth. The prompt also
encourages use of the full range, intermediate odd scores between even
anchors, and distinct scores for distinguishable videos. The latter
instruction may affect tie rates.

\subsection{Parsing, Normalization, and Paired Comparisons}
\label{app:vlm_aggregation}

Qwen3-VL and GPT-5.5 each produce 400 compliant responses. InternVL3 requires
offline formatting recovery for 26.5\% of responses; scores are extracted
from the original output without re-judging. One metric in one truncated
InternVL3 response remains missing and is excluded rather than scored zero.

For a valid raw score $s_{e,m,j,k}\in\{1,\ldots,10\}$ for episode $e$,
method $m$, judge $j$, and criterion $k$, the normalized score is
\begin{equation}
    q_{e,m,j,k}=s_{e,m,j,k}/10.
    \label{eq:vlm_normalization}
\end{equation}
Normalized values range from 0.1 to 1.0; raw integers are retained.
We report means and standard deviations by method, judge, and criterion,
with comparisons paired by episode and effective sample counts stated
where values are missing. All 12 judge--criterion aggregate differences
favor \method{}, with Cohen's $d$ from 0.36 to 1.37. These are standardized
method effects, distinct from the inter-rater agreement measured below.

\input{Appendix/appendix_human_vlm}

\subsection{Evaluation Source and Reproducibility}
\label{app:vlm_provenance}

We adapt the VLM-based evaluation pipeline of
WorldArena~\citep{shang2026worldarena} to the three judges and shared rubric
described above. The evaluation implementation is publicly available.
Per-video scores and original judge responses are retained for
verification, including responses that required formatting recovery.

\subsection{Limitations of VLM Evaluation}
\label{app:vlm_limitations}

Sparse frames can miss transient errors or late completion, particularly
in longer episodes. Shared evidence and rubric also make the judges'
errors potentially correlated; the 12 positive differences are not
independent replications. The static penalty explicitly penalizes freezing,
a common Full Context failure, so Motion Naturalness measures more than
smoothness. Sensitivity tests retain the aggregate preference direction,
but the 80-pair human comparison offers only limited corroboration.
These scores assess visible plausibility and instruction compatibility,
not verified physics, closed-loop performance, or real-robot task success.

\clearpage
\subsection{Full VLM Evaluation Prompt}
\label{app:vlm_prompt}

The original shared prompt is reproduced below, with only line wrapping
and typography adjusted. \texttt{\{instruction\}} is replaced by the episode instruction.

\tcbinputlisting{
    enhanced jigsaw,
    breakable,
    listing only,
    listing file={Appendix/vlm_prompt_v3.txt},
    colback=DeepPurple!5!white,
    colframe=DeepPurple!30!white,
    colbacktitle=DeepPurple!85!black,
    coltitle=white,
    fonttitle=\rmfamily\bfseries\large,
    title={VLM Judging Prompt},
    title after break={VLM Judging Prompt (continued)},
    boxrule=0.6pt,
    arc=2mm,
    left=3mm,
    right=3mm,
    top=2.5mm,
    bottom=2.5mm,
    before skip=10pt,
    after skip=10pt,
    listing options={
        language={},
        basicstyle=\rmfamily\fontsize{9.5}{10.5}\selectfont\color{black!90},
        numbers=none,
        breaklines=true,
        breakatwhitespace=true,
        columns=fullflexible,
        keepspaces=true,
        showstringspaces=false,
        xleftmargin=0pt,
        xrightmargin=0pt,
        aboveskip=0pt,
        belowskip=0pt,
        literate={—}{{\textemdash}}1
    }
}

%% file: Appendix/appendix_human_vlm.tex
\subsection{Human--VLM Agreement}
\label{app:human_vlm_agreement}

\begin{figure}[ht]
  \centering
  \includegraphics[width=\textwidth]{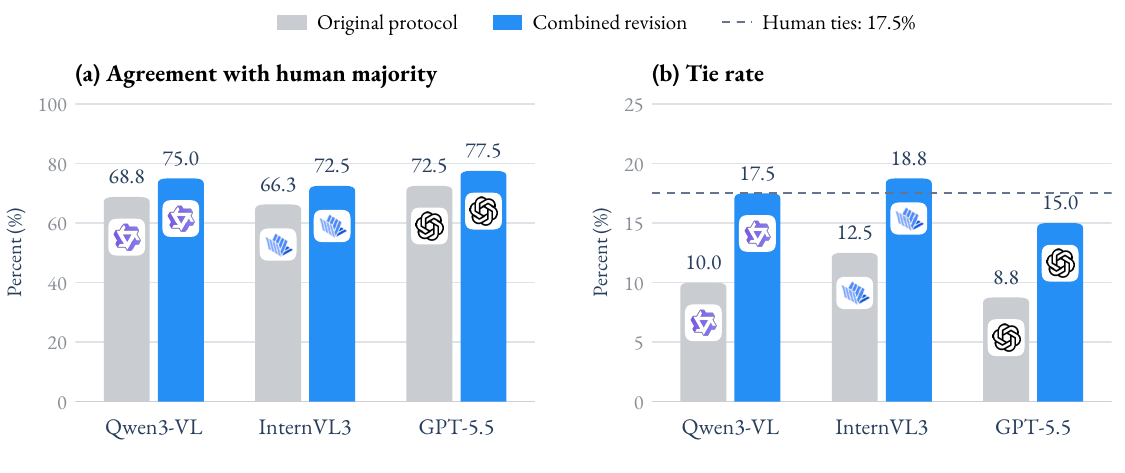}
  \caption{\textbf{Human--VLM alignment under two judging protocols.}
  Bars show pooled fractions on the same 80 pairs: (a) three-way agreement
  with human-majority labels and (b) tie rates. Gray denotes the original
  protocol and blue the combined revision. The dashed line in (b) marks
  the human tie rate of 17.5\%. Judge order is fixed across panels.}
  \label{fig:human_vlm_alignment}
\end{figure}

On 80 paired rollouts, human majorities favor \method{} in 42 pairs, tie
in 14, and favor Full Context in 24 (Table~\ref{tab:human_vlm_agreement}).
Original-protocol VLM agreement is 66.3--72.5\%, with unweighted
Cohen's $\kappa=0.43$--$0.52$. All judges produce fewer ties than humans
(Figure~\ref{fig:human_vlm_alignment}); this is consistent with, but does
not isolate, the prompt's discouragement of identical scores.

\begin{table}[ht]
  \centering
  \caption{\textbf{Human--VLM agreement on 80 paired rollouts.}
  Agreement counts all three labels, including ties. Unweighted Cohen's
  $\kappa$ is calculated from the pooled preference counts and observed
  agreement. The combined revision is described in
  Appendix~\ref{app:vlm_sensitivity}.}
  \label{tab:human_vlm_agreement}
  \small
  \setlength{\tabcolsep}{4pt}
  \renewcommand{\arraystretch}{1.15}
  \begin{tabular}{@{}llrrrcc@{}}
    \toprule
    Protocol & Evaluator & \shortstack{\method{}\\wins} & Ties
    & \shortstack{Full Context\\wins}
    & \shortstack{Agreement\\with humans} & $\kappa$ \\
    \midrule
    Reference & Human majority & 42 & 14 & 24 & -- & -- \\
    \midrule
    Original & \qwenjudge{} & 47 & 8 & 25 & 55/80 (68.8\%) & 0.46 \\
     & \internjudge{} & 45 & 10 & 25 & 53/80 (66.3\%) & 0.43 \\
     & \gptjudge{} & 49 & 7 & 24 & 58/80 (72.5\%) & 0.52 \\
    \midrule
    Combined & \qwenjudge{} & 43 & 14 & 23 & 60/80 (75.0\%) & 0.58 \\
     & \internjudge{} & 42 & 15 & 23 & 58/80 (72.5\%) & 0.55 \\
     & \gptjudge{} & 45 & 12 & 23 & 62/80 (77.5\%) & 0.62 \\
    \bottomrule
  \end{tabular}
\end{table}

The combined revision raises observed agreement to 72.5--77.5\% and
$\kappa$ to 0.55--0.62. These descriptive comparisons do not establish
statistical significance or a ranking of judge quality.

\FloatBarrier
\subsection{Sensitivity to the Judging Protocol}
\label{app:vlm_sensitivity}

Table~\ref{tab:vlm_sensitivity} compares protocol variants on the same 80
pairs. The combined revision allows equal scores, removes score-band
steering, includes both endpoint frames, and halves the static penalty.

\begin{table}[ht]
  \centering
  \caption{\textbf{Sensitivity of VLM preferences to the scoring protocol.}
  Each cell gives \method{} wins / ties / Full Context wins and sums to 80.
  The original protocol remains the basis of the main VLM results.}
  \label{tab:vlm_sensitivity}
  \small
  \setlength{\tabcolsep}{5pt}
  \renewcommand{\arraystretch}{1.15}
  \begin{tabular}{@{}lccc@{}}
    \toprule
    Protocol & \qwenjudge{} & \internjudge{} & \gptjudge{} \\
    \midrule
    Original & 47 / 8 / 25 & 45 / 10 / 25 & 49 / 7 / 24 \\
    Explicitly allow equal scores & 43 / 14 / 23 & 42 / 15 / 23 & 45 / 12 / 23 \\
    Half-strength static penalty & 45 / 11 / 24 & 43 / 13 / 24 & 47 / 10 / 23 \\
    Remove static penalty & 42 / 13 / 25 & 40 / 15 / 25 & 44 / 12 / 24 \\
    Include first and last frames & 46 / 11 / 23 & 44 / 13 / 23 & 48 / 9 / 23 \\
    Combined revision & 43 / 14 / 23 & 42 / 15 / 23 & 45 / 12 / 23 \\
    \bottomrule
  \end{tabular}
\end{table}

All tested judge--protocol combinations favor \method{} in aggregate,
including removal of the static penalty. Combined-revision tie rates
(15.0--18.8\%) approach the human rate (17.5\%). Allowing equal scores alone
produces the same marginal counts as the combined revision; this neither
implies identical per-pair labels nor isolates benefits from the other changes.

%% file: Appendix/appendix_prefix_length.tex
\section{Effect of Scene Prefix Length}
\label{app:prefix_length}

\subsection{Evaluation Setup and Token Allocation}
\label{app:prefix_setup}

We evaluate 450 held-out LIBERO-90 episodes with five generation seeds using
the final 20k checkpoint (Appendix~\ref{app:libero-reproducibility}) and a
1,931-token budget cap. Each configuration retains a scene prefix of
$p\in\{4,16,64,1280\}$ tokens, the initial 93-token visual--action block,
and a contiguous window of recent complete blocks
(Figure~\ref{fig:recap_token_sequence}). At steady state,

\begin{equation}
  W(p)=\left\lfloor\frac{1{,}931-p-93}{93}\right\rfloor,
  \qquad N(p)=p+93+93W(p).
  \label{eq:prefix_budget}
\end{equation}
The unused budget is $1{,}931-N(p)$: preserving complete blocks yields
slightly different effective lengths under the common cap.

\begin{table}[!ht]
  \centering
  \caption{\textbf{Scene-prefix allocation under a 1,931-token budget cap.}
  Results on 450 held-out LIBERO-90 episodes with five generation seeds are
  reported as mean $\pm$ standard deviation. Token counts describe the
  steady-state prompt; every row includes the same initial 93-token block.
  The full scene anchor is highlighted.}
  \label{tab:prefix_length}
  \footnotesize
  \renewcommand{\arraystretch}{1.2}
  \setlength{\tabcolsep}{3.2pt}
  \begin{tabular}{@{}rrrrrrcc@{}}
    \toprule
    \shortstack{Scene prefix\\(tokens)} &
    \shortstack{Initial block\\(tokens)} &
    \shortstack{Recent\\blocks} &
    \shortstack{Recent\\tokens} &
    \shortstack{Effective\\tokens} &
    \shortstack{Unused\\budget} &
    \shortstack{Late LPIPS\\$\downarrow$} &
    \shortstack{nDTW\\$\uparrow$} \\
    \midrule
    4 & 93 & 19 & 1,767 & 1,864 & 67 &
      $0.208\pm0.003$ & $0.619\pm0.006$ \\
    16 & 93 & 19 & 1,767 & 1,876 & 55 &
      $0.207\pm0.003$ & $0.621\pm0.006$ \\
    64 & 93 & 19 & 1,767 & 1,924 & 7 &
      $0.205\pm0.003$ & $0.624\pm0.005$ \\
    \rowcolor{DeepPurple!7}
    \textbf{1,280} & \textbf{93} & \textbf{6} & \textbf{558} &
      \textbf{1,931} & \textbf{0} &
      $\mathbf{0.203\pm0.002}$ & $\mathbf{0.628\pm0.005}$ \\
    \bottomrule
  \end{tabular}
\end{table}

\subsection{Results and Paired Comparisons}
\label{app:prefix_results}

The full scene anchor gives the best measured performance despite retaining
six recent blocks versus 19 for shorter prefixes (Table~\ref{tab:prefix_length}).
Relative to four scene tokens, it reduces Late LPIPS by 0.005 and increases
nDTW by 0.009. All six episode-level paired 95\% confidence intervals exclude
zero (Table~\ref{tab:prefix_paired}); gains over the 64-token prefix are modest.

\begin{table}[!ht]
  \centering
  \caption{\textbf{Episode-level paired comparisons.}
  Differences are full scene anchor minus the shorter scene prefix;
  brackets give 95\% confidence intervals. Lower Late LPIPS and higher nDTW
  favor the full anchor.}
  \label{tab:prefix_paired}
  \small
  \renewcommand{\arraystretch}{1.2}
  \setlength{\tabcolsep}{5pt}
  \begin{tabular}{@{}lcccc@{}}
    \toprule
    Comparison & $\Delta$ Late LPIPS $\downarrow$ & 95\% CI &
      $\Delta$ nDTW $\uparrow$ & 95\% CI \\
    \midrule
    Full anchor $-$ 4 tokens & $-0.005$ & $[-0.007,-0.003]$ &
      $+0.009$ & $[+0.005,+0.013]$ \\
    Full anchor $-$ 16 tokens & $-0.004$ & $[-0.006,-0.002]$ &
      $+0.007$ & $[+0.003,+0.011]$ \\
    Full anchor $-$ 64 tokens & $-0.002$ & $[-0.004,-0.001]$ &
      $+0.004$ & $[+0.001,+0.007]$ \\
    \bottomrule
  \end{tabular}
\end{table}

\FloatBarrier
\paragraph{Interpretation and scope.}
\label{app:prefix_scope}

This experiment tests joint allocation between scene context and recent
history, without isolating specific scene semantics. It complements the
LPIPS-AUC ablation in Figure~\ref{fig:rq3_mechanism}(a), where \method{} and
Recent Only show no measurable difference under that protocol. Implementations
retaining identical tokens, block boundaries, positions, and cache updates
realize the same schedule; this comparison distinguishes prefix allocations.

%% file: technical_report.bbl
\begin{thebibliography}{39}
\providecommand{\natexlab}[1]{#1}
\providecommand{\url}[1]{\texttt{#1}}
\expandafter\ifx\csname urlstyle\endcsname\relax
  \providecommand{\doi}[1]{doi: #1}\else
  \providecommand{\doi}{doi: \begingroup \urlstyle{rm}\Url}\fi

\bibitem[Bai et~al.(2025)Bai, Cai, Chen, Chen, Chen, Cheng, Deng, Ding, Gao, Ge, Ge, Guo, Huang, Huang, Huang, Hui, et~al.]{bai2025qwen3vl}
Shuai Bai, Yuxuan Cai, Ruizhe Chen, Keqin Chen, Xionghui Chen, Zesen Cheng, Lianghao Deng, Wei Ding, Chang Gao, Chunjiang Ge, Wenbin Ge, Zhifang Guo, Qidong Huang, Jie Huang, Fei Huang, Binyuan Hui, et~al.
\newblock {Qwen3-VL} technical report.
\newblock \emph{arXiv preprint arXiv:2511.21631}, 2025.

\bibitem[Barbero et~al.(2025)Barbero, Arroyo, Gu, Perivolaropoulos, Bronstein, Veli{\v{c}}kovi{\'c}, and Pascanu]{barbero2025llms}
Federico Barbero, {\'A}lvaro Arroyo, Xiangming Gu, Christos Perivolaropoulos, Michael Bronstein, Petar Veli{\v{c}}kovi{\'c}, and Razvan Pascanu.
\newblock Why do {LLMs} attend to the first token?
\newblock \emph{arXiv preprint arXiv:2504.02732}, 2025.

\bibitem[Bengio et~al.(2015)Bengio, Vinyals, Jaitly, and Shazeer]{bengio2015scheduled}
Samy Bengio, Oriol Vinyals, Navdeep Jaitly, and Noam Shazeer.
\newblock Scheduled sampling for sequence prediction with recurrent neural networks.
\newblock In \emph{Advances in Neural Information Processing Systems}, pages 1171--1179, 2015.

\bibitem[Brohan et~al.(2023)Brohan, Brown, Carbajal, Chebotar, Dabis, Finn, Gober, Hausman, Herzog, Hsu, et~al.]{brohan2023rt1}
Anthony Brohan, Noah Brown, Justice Carbajal, Yevgen Chebotar, Joseph Dabis, Chelsea Finn, Keerthana Gober, Karol Hausman, Alexander Herzog, Jasmine Hsu, et~al.
\newblock Rt-1: Robotics transformer for real-world control at scale.
\newblock \emph{arXiv preprint arXiv:2212.06817}, 2023.

\bibitem[Bruce et~al.(2024)Bruce, Dennis, Edwards, Parker-Holder, Shi, Hughes, Lai, Mavalankar, Steigerwald, Apps, et~al.]{bruce2024genie}
Jake Bruce, Michael~D Dennis, Ashley Edwards, Jack Parker-Holder, Yuge Shi, Edward Hughes, Matthew Lai, Aditi Mavalankar, Richie Steigerwald, Chris Apps, et~al.
\newblock Genie: Generative interactive environments.
\newblock In \emph{Forty-first International Conference on Machine Learning}, 2024.

\bibitem[Bulatov et~al.(2022)Bulatov, Kuratov, and Burtsev]{bulatov2022recurrent}
Aydar Bulatov, Yuri Kuratov, and Mikhail~S. Burtsev.
\newblock Recurrent memory transformer.
\newblock In \emph{Advances in Neural Information Processing Systems}, volume~35, 2022.

\bibitem[Dai et~al.(2019)Dai, Yang, Yang, Carbonell, Le, and Salakhutdinov]{dai2019transformerxl}
Zihang Dai, Zhilin Yang, Yiming Yang, Jaime Carbonell, Quoc~V. Le, and Ruslan Salakhutdinov.
\newblock {Transformer-XL}: Attentive language models beyond a fixed-length context.
\newblock In \emph{Proceedings of the 57th Annual Meeting of the Association for Computational Linguistics}, pages 2978--2988, Florence, Italy, 2019. Association for Computational Linguistics.
\newblock \doi{10.18653/v1/P19-1285}.

\bibitem[Ebert et~al.(2018)Ebert, Finn, Dasari, Xie, Lee, and Levine]{ebert2018visualforesight}
Frederik Ebert, Chelsea Finn, Sudeep Dasari, Annie Xie, Alex Lee, and Sergey Levine.
\newblock Visual foresight: Model-based deep reinforcement learning for vision-based robotic control.
\newblock In \emph{Robotics: Science and Systems}, 2018.

\bibitem[Esser et~al.(2021)Esser, Rombach, and Ommer]{esser2021taming}
Patrick Esser, Robin Rombach, and Bjorn Ommer.
\newblock Taming transformers for high-resolution image synthesis.
\newblock In \emph{Proceedings of the IEEE/CVF Conference on Computer Vision and Pattern Recognition}, pages 12873--12883, 2021.

\bibitem[Finn et~al.(2016)Finn, Goodfellow, and Levine]{finn2016unsupervised}
Chelsea Finn, Ian Goodfellow, and Sergey Levine.
\newblock Unsupervised learning for physical interaction through video prediction.
\newblock \emph{Advances in neural information processing systems}, 29, 2016.

\bibitem[Ge et~al.(2024)Ge, Zhang, Liu, Zhang, Han, and Gao]{ge2024fastgen}
Suyu Ge, Yunan Zhang, Liyuan Liu, Minjia Zhang, Jiawei Han, and Jianfeng Gao.
\newblock Model tells you what to discard: Adaptive {KV} cache compression for {LLMs}.
\newblock In \emph{International Conference on Learning Representations (ICLR)}, 2024.

\bibitem[Gu et~al.(2025{\natexlab{a}})Gu, Pang, Du, Liu, Zhang, Du, Wang, and Lin]{gu2025attentionsink}
Xiangming Gu, Tianyu Pang, Chao Du, Qian Liu, Fengzhuo Zhang, Cunxiao Du, Ye~Wang, and Min Lin.
\newblock When attention sink emerges in language models: An empirical view.
\newblock In \emph{International Conference on Learning Representations (ICLR)}, 2025{\natexlab{a}}.

\bibitem[Gu et~al.(2025{\natexlab{b}})Gu, Mao, and Shou]{gu2025longcontext}
Yuchao Gu, Weijia Mao, and Mike~Zheng Shou.
\newblock Long-context autoregressive video modeling with next-frame prediction.
\newblock \emph{arXiv preprint arXiv:2503.19325}, 2025{\natexlab{b}}.

\bibitem[Ha and Schmidhuber(2018)]{ha2018worldmodels}
David Ha and J{\"u}rgen Schmidhuber.
\newblock World models.
\newblock \emph{arXiv preprint arXiv:1803.10122}, 2018.

\bibitem[Hafner et~al.(2023)Hafner, Pasukonis, Ba, and Lillicrap]{hafner2023dreamerv3}
Danijar Hafner, Jurgis Pasukonis, Jimmy Ba, and Timothy Lillicrap.
\newblock Mastering diverse domains through world models.
\newblock \emph{arXiv preprint arXiv:2301.04104}, 2023.

\bibitem[Huang et~al.(2025)Huang, Li, He, Zhou, and Shechtman]{huang2025selfforcing}
Xun Huang, Zhengqi Li, Guande He, Mingyuan Zhou, and Eli Shechtman.
\newblock Self forcing: Bridging the train-test gap in autoregressive video diffusion.
\newblock In \emph{Advances in Neural Information Processing Systems}, 2025.

\bibitem[Li et~al.(2024)Li, Huang, Yang, Venkitesh, Locatelli, Ye, Cai, Lewis, and Chen]{li2024snapkv}
Yuhong Li, Yingbing Huang, Bowen Yang, Bharat Venkitesh, Acyr Locatelli, Hanchen Ye, Tianle Cai, Patrick Lewis, and Deming Chen.
\newblock {SnapKV}: {LLM} knows what you are looking for before generation.
\newblock In \emph{Advances in Neural Information Processing Systems (NeurIPS)}, 2024.

\bibitem[Liu et~al.(2024)Liu, Zhu, Gao, Feng, Liu, Zhu, and Stone]{liu2024libero}
Bo~Liu, Yifeng Zhu, Chongkai Gao, Yizhou Feng, Qiang Liu, Yuke Zhu, and Peter Stone.
\newblock {LIBERO}: Benchmarking knowledge transfer for lifelong robot learning.
\newblock In \emph{Advances in Neural Information Processing Systems}, volume~36, 2024.

\bibitem[Liu et~al.(2023)Liu, Desai, Liao, Wang, Xie, Xu, Kyrillidis, and Shrivastava]{liu2023scissorhands}
Zichang Liu, Aditya Desai, Fangshuo Liao, Weitao Wang, Victor Xie, Zhaozhuo Xu, Anastasios Kyrillidis, and Anshumali Shrivastava.
\newblock Scissorhands: Exploiting the persistence of importance hypothesis for {LLM} {KV} cache compression at test time.
\newblock In \emph{Advances in Neural Information Processing Systems (NeurIPS)}, 2023.

\bibitem[Mees et~al.(2022)Mees, Hermann, Rosete-Beas, and Burgard]{mees2022calvin}
Oier Mees, Lukas Hermann, Erick Rosete-Beas, and Wolfram Burgard.
\newblock {CALVIN}: A benchmark for language-conditioned policy learning for long-horizon robot manipulation tasks.
\newblock In \emph{IEEE Robotics and Automation Letters}, volume~7, pages 7327--7334, 2022.

\bibitem[Mentzer et~al.(2024)Mentzer, Minnen, Agustsson, and Toderici]{mentzer2024finite}
Fabian Mentzer, David Minnen, Eirikur Agustsson, and George Toderici.
\newblock Finite scalar quantization: {VQ-VAE} made simple.
\newblock In \emph{International Conference on Learning Representations}, 2024.

\bibitem[Micheli et~al.(2023)Micheli, Alonso, and Fleuret]{micheli2023iris}
Vincent Micheli, Eloi Alonso, and Fran{\c{c}}ois Fleuret.
\newblock Transformers are sample-efficient world models.
\newblock In \emph{International Conference on Learning Representations (ICLR)}, 2023.

\bibitem[{OpenAI}(2026)]{openai2026gpt55}
{OpenAI}.
\newblock {GPT-5.5} system card.
\newblock \url{https://openai.com/index/gpt-5-5-system-card/}, April 2026.
\newblock Accessed: 2026-08-06.

\bibitem[Po et~al.(2025)Po, Nitzan, Zhang, Chen, Dao, Shechtman, Wetzstein, and Huang]{po2025longcontext}
Ryan Po, Yotam Nitzan, Richard Zhang, Berlin Chen, Tri Dao, Eli Shechtman, Gordon Wetzstein, and Xun Huang.
\newblock Long-context state-space video world models.
\newblock In \emph{Proceedings of the IEEE/CVF International Conference on Computer Vision}, pages 8733--8744, 2025.

\bibitem[Rae et~al.(2020)Rae, Potapenko, Jayakumar, and Lillicrap]{rae2020compressive}
Jack~W. Rae, Anna Potapenko, Siddhant~M. Jayakumar, and Timothy~P. Lillicrap.
\newblock Compressive transformers for long-range sequence modelling.
\newblock In \emph{International Conference on Learning Representations}, 2020.

\bibitem[Ren et~al.(2025)Ren, Wei, Guo, Zhao, Kang, Feng, and Jin]{ren2025videoworld}
Zhongwei Ren, Yunchao Wei, Xun Guo, Yao Zhao, Bingyi Kang, Jiashi Feng, and Xiaojie Jin.
\newblock {VideoWorld}: Exploring knowledge learning from unlabeled videos.
\newblock In \emph{Proceedings of the IEEE/CVF Conference on Computer Vision and Pattern Recognition}, pages 29029--29039, 2025.

\bibitem[Shang et~al.(2026)Shang, Li, Ma, Su, Jin, Wang, Jin, Zhang, Tang, Su, et~al.]{shang2026worldarena}
Yu~Shang, Zhuohang Li, Yiding Ma, Weikang Su, Xin Jin, Ziyou Wang, Lei Jin, Xin Zhang, Yinzhou Tang, Haisheng Su, et~al.
\newblock {WorldArena}: A unified benchmark for evaluating perception and functional utility of embodied world models.
\newblock \emph{arXiv preprint arXiv:2602.08971}, 2026.

\bibitem[van~den Oord et~al.(2017)van~den Oord, Vinyals, and Kavukcuoglu]{oord2017neural}
Aaron van~den Oord, Oriol Vinyals, and Koray Kavukcuoglu.
\newblock Neural discrete representation learning.
\newblock In \emph{Advances in Neural Information Processing Systems}, 2017.

\bibitem[Vaswani et~al.(2017)Vaswani, Shazeer, Parmar, Uszkoreit, Jones, Gomez, Kaiser, and Polosukhin]{vaswani2017attention}
Ashish Vaswani, Noam Shazeer, Niki Parmar, Jakob Uszkoreit, Llion Jones, Aidan~N. Gomez, Lukasz Kaiser, and Illia Polosukhin.
\newblock Attention is all you need.
\newblock In \emph{Advances in Neural Information Processing Systems}, 2017.

\bibitem[Walke et~al.(2023)Walke, Black, Zhao, Vuong, Zheng, Hansen-Estruch, He, Myers, Kim, Du, et~al.]{walke2023bridgedata}
Homer Walke, Kevin Black, Tony~Z Zhao, Quan Vuong, Chongyi Zheng, Philippe Hansen-Estruch, Andre~Wang He, Vivek Myers, Moo~Jin Kim, Max Du, et~al.
\newblock Bridgedata v2: A dataset for robot learning at scale.
\newblock \emph{arXiv preprint arXiv:2308.12952}, 2023.

\bibitem[Wu et~al.(2026)Wu, Li, Gao, Xu, Zhang, Wang, Zhao, Wang, Zhao, Wang, et~al.]{wu2026roboalign}
Hao Wu, Yuqi Li, Yuan Gao, Fan Xu, Fan Zhang, Kun Wang, Penghao Zhao, Qiufeng Wang, Yizhou Zhao, Weiyan Wang, et~al.
\newblock Roboalign-r1: Distilled multimodal reward alignment for robot video world models.
\newblock \emph{arXiv preprint arXiv:2605.03821}, 2026.

\bibitem[Wu et~al.(2024)Wu, Yin, Feng, He, Li, Hao, and Long]{wu2024ivideogpt}
Jialong Wu, Shaofeng Yin, Ningya Feng, Xu~He, Dong Li, Jianye Hao, and Mingsheng Long.
\newblock ivideogpt: Interactive videogpts are scalable world models.
\newblock \emph{Advances in Neural Information Processing Systems}, 37:\penalty0 68082--68119, 2024.

\bibitem[Wu et~al.(2025)Wu, Yin, Feng, and Long]{wu2025rlvr}
Jialong Wu, Shaofeng Yin, Ningya Feng, and Mingsheng Long.
\newblock Rlvr-world: Training world models with reinforcement learning.
\newblock \emph{arXiv preprint arXiv:2505.13934}, 2025.

\bibitem[Xiao et~al.(2024)Xiao, Tian, Chen, Han, and Lewis]{xiao2024streamingllm}
Guangxuan Xiao, Yuandong Tian, Beidi Chen, Song Han, and Mike Lewis.
\newblock Efficient streaming language models with attention sinks.
\newblock In \emph{International Conference on Learning Representations (ICLR)}, 2024.

\bibitem[Yan et~al.(2021)Yan, Zhang, Abbeel, and Srinivas]{yan2021videogpt}
Wilson Yan, Yunzhi Zhang, Pieter Abbeel, and Aravind Srinivas.
\newblock Videogpt: Video generation using vq-vae and transformers.
\newblock In \emph{Advances in Neural Information Processing Systems}, 2021.

\bibitem[Zhang et~al.(2025)Zhang, Cai, Li, Wetzstein, and Agrawala]{zhang2025framepack}
Lvmin Zhang, Shengqu Cai, Muyang Li, Gordon Wetzstein, and Maneesh Agrawala.
\newblock Frame context packing and drift prevention in next-frame-prediction video diffusion models.
\newblock In \emph{Advances in Neural Information Processing Systems}, 2025.

\bibitem[Zhang et~al.(2023)Zhang, Sheng, Zhou, Chen, Zheng, Cai, Song, Tian, R{\'e}, Barrett, Wang, and Chen]{zhang2023h2o}
Zhenyu Zhang, Ying Sheng, Tianyi Zhou, Tianlong Chen, Lianmin Zheng, Ruisi Cai, Zhao Song, Yuandong Tian, Christopher R{\'e}, Clark Barrett, Zhangyang Wang, and Beidi Chen.
\newblock {H2O}: Heavy-hitter oracle for efficient generative inference of large language models.
\newblock In \emph{Advances in Neural Information Processing Systems (NeurIPS)}, 2023.

\bibitem[Zhou et~al.(2024)Zhou, Du, Chen, Li, Yeung, and Gan]{zhou2024robodreamer}
Siyuan Zhou, Yilun Du, Jiaben Chen, Yandong Li, Dit-Yan Yeung, and Chuang Gan.
\newblock Robodreamer: Learning compositional world models for robot imagination.
\newblock \emph{arXiv preprint arXiv:2404.12377}, 2024.

\bibitem[Zhu et~al.(2025)Zhu, Wang, Chen, Liu, Ye, Gu, Tian, Duan, Su, Shao, Gao, Cui, et~al.]{zhu2025internvl3}
Jinguo Zhu, Weiyun Wang, Zhe Chen, Zhaoyang Liu, Shenglong Ye, Lixin Gu, Hao Tian, Yuchen Duan, Weijie Su, Jie Shao, Zhangwei Gao, Erfei Cui, et~al.
\newblock {InternVL3}: Exploring advanced training and test-time recipes for open-source multimodal models.
\newblock \emph{arXiv preprint arXiv:2504.10479}, 2025.

\end{thebibliography}
